\documentclass[alpha-refs]{wiley-article}

\usepackage{graphicx}
\usepackage[space]{grffile}
\usepackage{latexsym}
\usepackage{textcomp}
\usepackage{longtable}
\usepackage{tabulary}
\usepackage{booktabs,array,multirow}
\usepackage{amsfonts,amsmath,amssymb}
\usepackage{dsfont}
\usepackage{natbib}
\usepackage{url}
\usepackage{hyperref}
\hypersetup{colorlinks=false,pdfborder={0 0 0}}
\usepackage{etoolbox}
\graphicspath{{figures/}{figures/Figures/}}

\AtBeginDocument{\DeclareGraphicsExtensions{.pdf,.PDF,.eps,.EPS,.png,.PNG,.tif,.TIF,.jpg,.JPG,.jpeg,.JPEG}}

\usepackage[utf8]{inputenc}
\usepackage[english]{babel}

\usepackage{siunitx}

\paperfield{arXiv}

\corraddress{Catherine George, Mathematisches Institut für Maschinelles Lernen und Data Science, Katholische Universität Eichstätt-Ingolstadt, Ingolstadt, Bavaria, 85049, Germany}
\corremail{catherine.george@ku.de}
\fundinginfo{Klaus Tschira Stiftung gGmbH Projekt-Nr./FKZ: 00.019.2024: "Uncertainty-aware and physics-informed machine learning for short-range atmospheric forecasts".}

\papertype{Original Article}

\title{Uncertainty quantification via conformal prediction in data assimilation}

\author[1]{Catherine George}
\author[2]{Alireza Javanmardi}
\author[3]{Tijana Janji\'{c}}
\author[4]{Eyke Hüllermeier}

\affil[1,3]{Mathematisches Institut für Maschinelles Lernen und Data Science, Katholische Universität Eichstätt-Ingolstadt}

\affil[2,4]{Institute of Informatics, LMU Munich\\Munich Center for Machine Learning (MCML)}

\runningauthor{C.George et al.}
\begin{document}

\maketitle
\begin{abstract}
Quantifying the evolution of uncertainty is critical to both probabilistic forecasting and data assimilation in numerical weather prediction. In this study, we investigate the applicability of conformal prediction (CP), a recent machine learning (ML) method, to quantify uncertainty in a controlled, idealized setting. We use the one dimensional modified shallow water model, designed to mimic the convective process. CP provides a set of possible outcomes with a chosen confidence level. Here, we compare and evaluate the average empirical coverage, the average interval length, miss low, miss high and average interval score loss (AISL) for three variants of CP, namely a) Standard CP, b) Normalized CP and c) Conformalized Quantile Regression. We further compare these CP-based uncertainty estimates with traditional ensemble-based measures such as standard deviation intervals and ensemble spread. In addition, we investigate the integration of CP-derived uncertainty within the data assimilation cycle through CP perturbations. Our results highlight the strengths and limitations of each approach, providing insight into the effectiveness of CP to complement common ensemble-based uncertainty quantification in simplified atmospheric models.

\end{abstract}
\section{Introduction}

{\label{sec:Introduction}}
Uncertainty is an inseparable part of atmospheric prediction. Due to the chaotic dynamics of the atmosphere, the limitations of numerical models, and the incompleteness of observations, no single forecast can ever be fully certain. Quantifying this uncertainty is crucial for informed decision-making \citep{LEUTBECHER20083515}, especially in high-impact weather events where confidence intervals can be as important as the forecast itself. In operational meteorology, uncertainty estimation plays a key role in risk assessment, early warning systems, and decision support for extreme events. In numerical weather prediction, uncertainty estimates are also required in data assimilation, which combines information from numerical models and observations to approximate the most likely atmospheric state \cite{Kalnay_2002, EnKFbook}. Data assimilation methods aim to optimally merge prior model forecasts with observations while accounting for uncertainties in both sources of information. The uncertainty arises from several components, including errors in the background forecast, observational errors, and model imperfections. Accurately quantifying these uncertainties is essential, as they determine the relative weighting of model forecasts and observations during the assimilation process. 

Among data assimilation approaches, ensemble-based data assimilation methods such as the ensemble Kalman filter (EnKF) represent uncertainty by evolving multiple realizations of the atmospheric state \citep{E94, Burgers98}. In these methods, uncertainty in the background forecast is characterized through the ensemble spread, which provides a sample-based estimate of the forecast error covariance \citep{Evensen2003}. Traditionally, ensemble methods have therefore been the cornerstone of uncertainty quantification in numerical weather prediction \cite{LEUTBECHER20083515}. By evolving multiple model realizations from perturbed initial states, ensembles approximate the distribution of possible outcomes \citep{Magnusson181520}. Despite their success, ensemble-based approaches face important challenges. In particular, reliable estimation of forecasting uncertainty requires sufficiently large ensembles, which makes these methods computationally demanding \citep{CHATTOPADHYAY2023111918,Janjicetal_2023}. Moreover, ensemble sizes are often limited in operational settings due to computational constraints, which can lead to sampling errors and underestimation of uncertainty in the estimated error covariances \citep{EnKFbook}. These challenges motivate the exploration of alternative approaches for uncertainty quantification.

Motivated by the computational expensiveness and slow convergence of Monte Carlo based ensemble methods in high-dimensional systems, uncertainty-aware machine learning (ML) is emerging as a promising alternative or complement to ensembles \citep{Price2025}. ML methods, particularly deep neural networks, are powerful function approximators capable of learning complex nonlinear relationships directly from data \citep{LeCun,Goodfellow-et-al-2016,LeglerJanjic22}. In recent years, ML has increasingly been explored in meteorology and climate science for tasks such as parameterization and forecasting \citep{article,doi:10.1073/pnas.1810286115}. Beyond improving predictive accuracy, several ML approaches aim to explicitly represent predictive uncertainty. Many ML methods quantify uncertainty through prediction intervals or probabilistic forecasts, which describe a range of plausible future states rather than a single deterministic prediction. For instance, Bayesian neural networks (BNNs) model uncertainty in network weights \citep{neal_bl,Goan_2020}, while deep ensembles use variability across independently trained models as a proxy for model uncertainty \citep{laks_sa17}. Other approaches, such as dropout-based Bayesian approximations \citep{gal_da16} or random forests \citep{brei_rf01}, can also produce probabilistic predictions. These methods provide valuable insights into predictive uncertainty but typically rely on heuristic assumptions or approximate inference schemes. However, ML-based uncertainty quantification is not without challenges. Many probabilistic predictions are miscalibrated, systematically over or underestimating uncertainty \cite{guo17a, ImanolJMLR:v23:22-0658}. As a result, prediction intervals may fail to contain the true state at the desired confidence level. In other words, most ML models lack coverage guarantees, meaning that the probability that the true value lies within the predicted interval is not controlled. In applications such as short-range weather forecasting and data assimilation, where reliable uncertainty estimates are essential, this limitation can lead to misleading confidence in predictions.

To overcome this gap, conformal prediction (CP) offers a statistically rigorous and distribution-free framework for uncertainty quantification \cite{angelopoulos_bates_2022}. CP produces prediction intervals with guaranteed finite sample coverage, which means that the true value lies within the set at a user-specified confidence level. Unlike ensemble spread or heuristic ML uncertainty estimates, CP provides formal reliability guarantees under very mild assumptions. Recent work has demonstrated the potential of CP for uncertainty quantification in machine learning applications \citep{romano2019conformalized, javanmardi2023conformal}. Because conformal prediction does not assume a particular data distribution, it is particularly attractive for atmospheric applications where distributions are often complex and non-Gaussian. To test and develop new uncertainty quantification methods, researchers often use toy models. In our study, we use the one dimensional modified shallow water (MSW) model of \citet{wuersch_craig_2014}, later employed by \citet{Haslehneretal_2016p2} and \citet{Ruckstuhletal_2021}, for the study of data assimilation methods. The model captures essential convective scale dynamics while remaining computationally affordable. The MSW model reproduces regime switches and rainfall onset delays that resemble convection, making it an ideal benchmark for uncertainty-aware methods. Previous work has demonstrated the importance of enforcing physical constraints in data assimilation \cite{Janjicetal_2014,Zenghttps://doi.org/10.1002/qj.3142,Ruckstuhl_Janjic_2018} and explored ML-data assimilation hybrids such as convolution neural network-based corrections to conserve mass \cite{Ruckstuhletal_2021}. Building on this foundation, the present study introduces CP as a novel post-hoc uncertainty quantification tool for the MSW model. Our contributions are: (i) We adapt CP to quantify uncertainty in the spatio-temporal setting of a convective scale toy model, (ii) we employ different CP algorithms and evaluate and compare their coverage, interval sharpness, miss low, miss high rates and average interval score loss, (iii) similarly we compare the different CP methods with traditional uncertainty quantification methods of standard deviation interval and ensemble spread and (iv) we explore the potential of CP to improve data assimilation by introducing CP perturbations either through the convolutional neural network (CNN) output prior to assimilation or directly within the EnKF analysis. The paper is organized as follows. Section 2 introduces the MSW model, data assimilation, training and architecture of the CNN. Section 3 covers uncertainty quantification by CP and describes CP variants along with its experimental setup. Section 4 presents the approach for incorporating  CP in data assimilation to improve it. Section 5 discusses the results and Section 6 concludes the paper.

\section{Experimental Setup}
\subsection{Modified Shallow Water Model}
{\label{sec:model}}
In this study, we use the one dimensional MSW model \cite{wuersch_craig_2014}, adopting the numerical implementation as proposed therein. The model captures the dynamics of horizontal wind velocity 'u', fluid height 'h' and rain water content 'r' and has been used as an idealized framework for studying data assimilation algorithms on convective scale \cite{Ruckstuhl_Janjic_2018,Janjicetal_2021}. Its design allows for the convection representation through simple threshold based triggering and localized perturbations. It ensures conservation of mass, i.e., the spatial integral of fluid height over the domain is constant in time, and the rain is nonnegative. If rain becomes negative during model integration, negative values are set to zero. We adopt the configuration used in the earlier studies \cite{Ruckstuhletal_2021}, with a 125 km one-dimensional domain discretized into 250 grid points and integrated forward in time with a step size of 5 seconds. The MSW model equations are given by:

\begin{equation}\label{eq:u}
\frac{\partial u}{\partial t} + u\frac{\partial u}{\partial x}+\frac{\partial (\phi + \gamma^2 r)}{\partial x} = \beta_u + D_u\frac{\partial^2 u}{\partial x^2}, 
\end{equation}
with 
\begin{equation} 
\phi = \left\{ 
\begin{array}{ll} \phi_c  & \text{if } h > h_c\\
                            gh  &\text{otherwise,}
\end{array}
\right.
\end{equation}
\begin{equation} 
\frac{\partial r}{\partial t} + u\frac{\partial r}{\partial x}
= D_r\frac{\partial^2 r}{\partial x^2} -\eta r - 
\left\{ 
\begin{array}{ll} \delta \frac{\partial u}{\partial x}, &\text{if} h>h_r \text{and} \frac{\partial u}{\partial x} <0 \\
                            0  &\text{otherwise,}
\end{array}
\right.  
\end{equation}
\begin{equation}
\frac{\partial h}{\partial t} +
\frac{\partial (uh)}{\partial x} = D_h \frac{\partial^2 h}{\partial x^2}.
\end{equation}

The parameters that control the switch between regimes are fixed and have the following values
$ \phi_c = 899.77 \, m^{2}s^{-2}, \, \,  h_c = 90.02  \, m, \, \, h_r = 90.4  \, m$. In the equation for the evolution of rain, the parameter $\eta=2.5 \cdot 10^{-4} $ is the removal rate of the rain while $\delta=1/300$ is the production rate. The diffusion parameters are given by $ D_u = D_h =25000  \, m^{2} s^{-1}, D_r = 200  \, m^{2} s^{-1}$ for $u,h$ and $r$, respectively. The absolute fluid level is $h_0 = 90 \,m$ and $\gamma=\sqrt{gh_0}$ is the gravity wave speed where $g$ is the gravity constant.
To initiate convection, stochastic Gaussian forcing $\beta_u$ is added at random locations in each model time step. The $\beta_u$ has a half width of four grid points and an amplitude of 0.002 \(m/s^2\). 


\subsection{Data assimilation and training the CNN}
{\label{sec:Data}}
The experimental setup follows a twin-experiment design.The MSW model run will represent the true state of the system, and we call it the nature run. We generate synthetic observations from a nature run by perturbing the nature run with Gaussian errors for wind velocity and fluid height, and log-normal errors for precipitation. In addition, observations are taken only at grid points where precipitation exceeds $0.005$ dBZ, with an additional $10\%$ of grid points containing wind data to mimic supplementary measurements. It should be noted that the nature run is used solely for the generation of observations and the evaluation of results and is not involved in any part of the data assimilation processes. Two data assimilation methods are used to create analysis: the ensemble Kalman filter \citep[EnKF;][]{E94,Burgers98,EnKFbook} and quadratic programming ensemble \citep[QPEns;][]{Janjicetal_2014} that preserves mass and non-negativity of rain. 
The EnKF produces an ensemble of analyses that are propagated forward in time during each assimilation cycle using the MSW model. This process yields a background ensemble at fixed time $k$ designed to represent the uncertainty in the model forecasts. The stochastic EnKF uses this ensemble of predictions (backgrounds) $\{\textbf{Z}^{b,i}\}_{i=1}^N$ to calculate the sample background error covariance matrix 
\begin{equation}
\textbf{P}^b = \frac{1}{N-1} \sum_{i=1}^{N} [\textbf{Z}^{b,i}-\bar{\textbf{Z}}^b][\textbf{Z}^{b,i}- \bar{\textbf{Z}}^b]^T 
\end{equation}
with $\bar{\textbf{Z}}^b$ representing the ensemble mean.  
The EnKF then applies the standard Kalman filter equations to compute the analysis for each ensemble member $i=1,\ldots,N$ using $\textbf{P}^b$ and observation error covariance $\textbf{R}$.  The QPEns does not use Kalman filter equations; instead, it minimizes the cost function with additional physical constraints in each ensemble member and uses $\textbf{P}^b$ and $\textbf{R}$ to weight the background and observations, respectively. It is shown to be more accurate than EnKF for state estimation, due to the explicit additional use of physical information \cite{Janjicetal_2014, Zenghttps://doi.org/10.1002/qj.3142, Ruckstuhl_Janjic_2018, gleiteretal_2022}. However, QPEns is computationally expensive compared to EnKF, which motivated \citet{Ruckstuhletal_2021} to train a convolutional neural network (CNN) to map EnKF analysis to QPEns analysis, thereby achieving QPEns-like quality at a fraction of the computational cost. Both QPEns and EnKF produce an ensemble of background and analysis in each data assimilation cycle. Following \citet{Ruckstuhletal_2021}, using an ensemble of $N=10$ members initialized from arbitrary states, we start the data assimilation cycling and train a CNN that inputs EnKF analysis ensemble and outputs QPEns analysis ensemble. Also, the CNN is not a standard black-box network; it incorporates a mass conservation constraint to ensure that the physical consistency is preserved.  We generate training data using 10 random seeds, consisting of QPEns cycling outputs {$Q_t^b$,$Q_t^a$ : t = 1,2,...,200}, where Q denotes the QPEns system, and the superscripts b and a represent the background and analysis states, respectively. Simultaneously, we construct the dataset {$X_t^a$ : t = 1,2,...,200}, where $X_t^a$ denotes the unconstrained solution obtained from the same background $Q_t^b$. For each random seed, the data lies in $\mathbb{R}^{N \times n \times 3}$, where $N$ is the number of ensemble members, n is the number of grid points ($n=250$), and the three channels correspond to the variables u, h, and r, over 200 assimilation cycles. The corresponding output data has the same dimensional structure, with the CNN applied to the EnKF analysis for the subsequent 180 assimilation cycles, starting from timestep t = 20.

\begin{figure}[h!]
\begin{center}
\includegraphics[width=0.70\columnwidth]{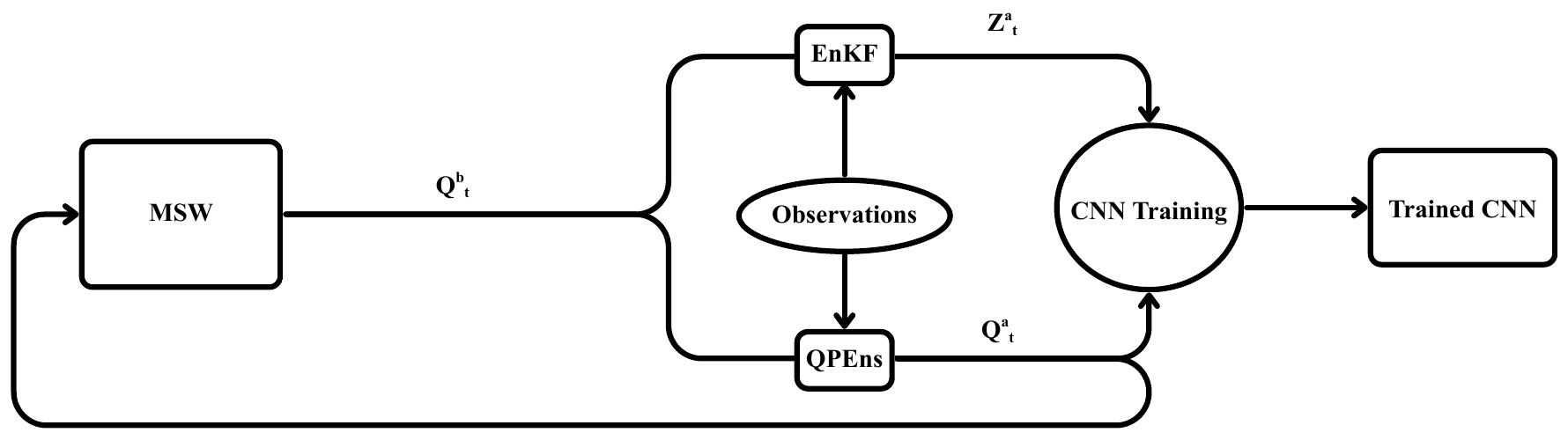}
\caption{Schematic overview of the CNN training workflow within the data assimilation framework.}
\label{Training}
\end{center}
\end{figure}

The CNN is trained to learn the correction introduced by the constrained QPEns analysis relative to the unconstrained EnKF update. To prevent the influence of large transient increments associated with the initial spin-up phase, the first 20 assimilation cycles are excluded from the training dataset. QPEns cycling is performed initially, and only states obtained from cycle 20 onward, when the system has reached a dynamically balanced state, are used to construct the training pairs. To generate the training data set, QPEns cycling produces background states $Q_t^b$ and corresponding constrained analyses $(Q_t^a)$. At each assimilation timestep, the same QPEns background $(Q_t^b)$ is used to compute an unconstrained EnKF analysis $(Z_t^a)$ by applying the standard Kalman update without enforcing physical constraints. The CNN is then trained to map the unconstrained analysis $(Z_t^a)$ to the constrained QPEns analysis $(Q_t^a)$. By using identical background states for both updates, the training procedure isolates the difference between the unconstrained and constrained minimization steps, ensuring that the CNN learns only the effect of constraint enforcement, rather than differences caused by accumulated background errors during cycling. Wind velocity '$u$' and fluid height '$h$' are normalized by subtracting their climatological mean and dividing by the standard deviation, while precipitation '$r$' is scaled only by its standard deviation to preserve positivity. The overall CNN training pipeline adopted in this study is illustrated schematically in \autoref{Training}. For further details, we refer to \citet{Ruckstuhletal_2021}.

For the evaluation experiments, QPEns is first integrated independently for 20 assimilation cycles to allow the system to reach a dynamically balanced state. The analysis ensemble obtained at cycle 20 is then used as a common initial condition for the EnKF and CNN corrected EnKF experiments. From cycle 21 onward, three independent data assimilation systems are run: EnKF, QPEns, and EnKF augmented with the trained CNN correction. Each method subsequently performs its own forecast and analysis steps without further interaction, ensuring a consistent and fair comparison of their respective performances.

\subsection{Architecture of the CNN}
{\label{sec:CNN}}
To emulate the relationship between the unconstrained and physically constrained analyses, CNN is employed. As in \citet{Ruckstuhletal_2021}, the CNN consists of four convolutional hidden layers, each containing 32 filters with a kernel size of 3. The activation function used in all hidden layers is the scaled exponential linear unit (SELU), which helps maintain normalized activations throughout the network and improves training stability. The output layer is also a convolutional layer, with three filters corresponding to the three model variables - u, h, r and a kernel size of 3.
A linear activation is applied to the output channels for u and h, while a ReLU activation ensured non-negative predictions for r. The effective influence radius of the CNN, is five grid points. Training is conducted using the Adam optimizer, and the loss function is defined as the root mean square error (RMSE) between the predicted and target analyses, computed over all grid points and averaged over the three variables u, h, and r.
To prevent overfitting and monitor generalization, a validation dataset is used alongside the training data, and the model is trained for 100 epochs with a batch size of 96. The implementation is carried out in Python using Keras (Chollet, 2017) and is based on the implementation of \citet{Ruckstuhletal_2021}.

\section{Uncertainty quantification via CP}
\begin{figure}[h!]
\begin{center}
\includegraphics[width=0.70\columnwidth]{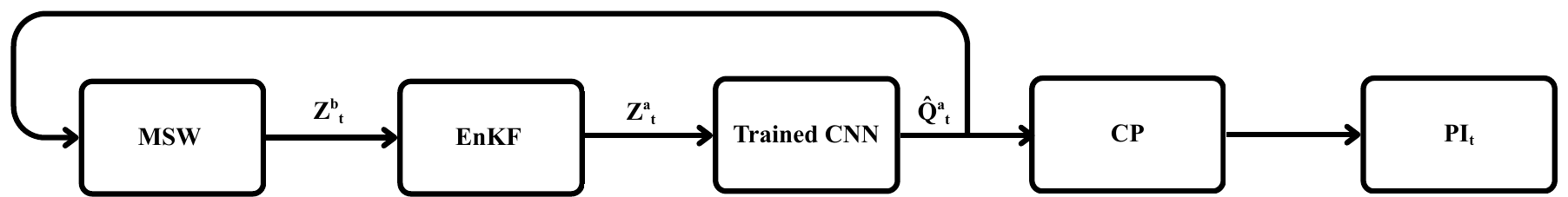}
\caption{CP pipeline for generating prediction intervals from CNN analysis.}
\label{CP}
\end{center}
\end{figure}

{\label{sec:UQ}} 
In the system described above, the CNN takes the EnKF analysis $Z_t^{a, \text{new}}$ as input and outputs an estimate $\hat{Q}_t^{a, \text{new}}$ of the QPEns analysis $Q_t^{a, \text{new}}$\footnote{With a slight abuse of notation, we treat $Z_t^{a,i}$ and $Q_t^{a,i}$ (and similarly $Z_t^{a,\text{new}}$ and $Q_t^{a,\text{new}}$) as scalar quantities corresponding to a single variable (among $r, u, h$) at assimilation time $t$. The indexing (e.g., $i$ or “new”) is used generically and may refer to any instance among the $N \times n$ instances (e.g., across ensemble members or spatial grid points).}. To equip this prediction with a reliable measure of uncertainty, we apply the CP framework to construct a distribution-free prediction interval $\mathrm{PI}(Z_t^{a, \text{new}}, \alpha) \subset \mathbb{R}$ that contains the sought variable $Q_t^{a, \text{new}}$ with a high probability $1-\alpha$, where $\alpha \in (0,1)$ is a user-defined error rate, as illustrated in \autoref{CP}. In this context, uncertainty is represented by the length of the $\mathrm{PI}$; a wider interval indicates higher uncertainty. 
At its core, CP performs hypothesis testing. Given an input $Z_t^{a, \text{new}}$, it evaluates every candidate output $y \in \mathbb{R}$ to determine if it could plausibly be the valid label. A candidate is only excluded if the hypothesis is rejected. Hypothesis testing is performed by calculating a nonconformity score, a real-valued function $S(Z_t^{a, \text{new}}, y)$ that quantifies how "unusual" the pair $(Z_t^{a, \text{new}}, y)$ appears relative to previously observed data. In this study, the QPEns analysis is used as a proxy truth for the evaluation of the CP intervals.

There are two main variants of conformal prediction: Full CP and Split CP.  Full CP is a transductive method, in which the nonconformity score of each candidate label $y$ is computed by retraining the model on the full dataset augmented with the test point $(Z_t^{a, \text{new}}, y)$, and comparing it against the scores of all other data points. While this yields tight prediction intervals, it is computationally prohibitive since it requires a full model retraining for every candidate $y$. Split CP, by contrast, is an inductive method that trains the model once and uses a separate held-out calibration set to estimate the nonconformity score distribution, making it computationally efficient and well-suited to settings where the model is expensive to train. For this reason, we focus on Split CP in this paper. Suppose we are given a dataset of paired observations $\{(Z_t^{a,i}, Q_t^{a,i})\}_{i=1}^M$. Split CP begins by randomly partitioning the data indices $[M]$ into two disjoint subsets: a training set $\mathcal{I}_{\mathrm{train}}$ and a calibration set $\mathcal{I}_{\mathrm{cal}}$. A CNN is trained exclusively on $\mathcal{I}_{\mathrm{train}}$ to produce point predictions $\hat{Q}_t^{a,i}$. The nonconformity score is then evaluated on the calibration data, yielding the set of scores $\{ S(Z_t^{a,i}, Q_t^{a,i}) : i \in \mathcal{I}_{\mathrm{cal}} \}$. We denote by $\hat{q}_{1-\alpha}$ the $(1-\alpha)(1 + 1/|\mathcal{I}_{\mathrm{cal}}|)$ quantile of this set of scores. The resulting prediction interval is defined as

\begin{align}
    \mathrm{PI}(Z_t^{a, \text{new}}, \alpha) = \{ y \in \mathbb{R} : S(Z_t^{a, \text{new}}, y) \le \hat q_{1-\alpha} \}.
\end{align}
If the data points $\{(Z_t^{a,i}, Q_t^{a,i})\}_{i \in \mathcal{I}_{\mathrm{cal}}} \cup (Z_t^{a, \text{new}}, Q_t^{a, \text{new}})$ are exchangeable, meaning their joint distribution is invariant under permutation, then the Split CP procedure described above satisfies the marginal coverage guarantee:
\begin{align}
    \mathbb{P}\big(Q_t^{a, \text{new}} \in \mathrm{PI}(Z_t^{a, \text{new}}, \alpha)\big) \ge 1 - \alpha.
\end{align}
This guarantee is distribution-free and holds regardless of the choice of nonconformity score. However, the nonconformity score significantly influences the length of the prediction intervals, a property referred to as efficiency, as well as the shape of the intervals, as we will clarify in the following sections. In this paper, we focus on three popular methods in CP: standard CP (SCP) \cite{vovk2022algorithmic}, normalized CP (NCP) \cite{papadopoulos2008normalized}, and conformalized quantile regression (CQR) \cite{romano2019conformalized}. Figure~\ref{CP} illustrates the proposed framework of integrating CP to the current setup to predict intervals.

\subsection{Standard CP}
The nonconformity score $S(Z_t^{a,i}, Q_t^{a,i})$ is often defined by the level of disagreement between the point prediction $\hat{Q}_t^{a,i}$ and the true value $Q_t^{a,i}$. Since these are real-valued numbers, a standard choice is the absolute residual:
\begin{equation}\label{eq:scores:sbsolute residual}
    S^{\text{SCP}}(Z_t^{a,i}, Q_t^{a,i}) = | \hat{Q}_t^{a,i} - Q_t^{a,i} |.
\end{equation}
Using this nonconformity score, the prediction interval $\mathrm{PI}(Z_t^{a,\text{new}}, \alpha)$ can be written as:
\begin{equation}\label{eq:intervals:sbsolute residual}
    \mathrm{PI}^{\text{SCP}}(Z_t^{a,\text{new}}, \alpha) = [\hat{Q}_t^{a,\text{new}} - \hat{q}_{1-\alpha}, \hat{Q}_t^{a,\text{new}} + \hat{q}_{1-\alpha}].
\end{equation}
An immediate drawback of this approach is that the prediction intervals have a constant length of $2 \hat{q}_{1-\alpha}$ for every point. As mentioned previously, the interval length represents the uncertainty of the prediction; therefore, one would expect the length to adapt to the difficulty of the prediction at each point. This constant-width behaviour fails to account for heteroscedasticity, where some regions of the input space are inherently more uncertain than others.
\subsection{Normalized CP}
To address this, more sophisticated nonconformity scores can be used to achieve adaptivity. One such method is normalized CP, which scales the residual by a heuristic uncertainty estimate $\hat{\sigma}(Z_t^{a,i})$:
\begin{equation}\label{eq:scores:normalized}
    S^{\text{NCP}}(Z_t^{a,i}, Q_t^{a,i}) = \frac{|\hat{Q}_t^{a,i} - Q_t^{a,i}|}{\hat{\sigma}(Z_t^{a,i})}.
\end{equation}
Accordingly, the resulting prediction interval becomes:
\begin{equation}\label{eq:intervals:normalized}
    \mathrm{PI}^{\text{NCP}}(Z_t^{a,\text{new}}, \alpha) = [\hat{Q}_t^{a,\text{new}} - \hat{q}_{1-\alpha} \hat{\sigma}(Z_t^{a,\text{new}}), \hat{Q}_t^{a,\text{new}} + \hat{q}_{1-\alpha} \hat{\sigma}(Z_t^{a,\text{new}})].
\end{equation}
In this case, the length of the prediction interval at each point $Z_t^{a,\text{new}}$ is proportional to the uncertainty estimate $\hat{\sigma}(Z_t^{a,\text{new}})$ at that point. A common approach is to define $\hat{\sigma}(Z_t^{a,\text{new}})$ such that it estimates the local dispersion of the residuals at $Z_t^{a,\text{new}}$. For this purpose, an auxiliary regressor can be trained on the training data $\{ (Z_t^{a,i}, |Q_t^{a,i} - \hat{Q}_t^{a,i}|) : i \in \mathcal{I}_{\mathrm{train}} \}$.
\subsection{CQR}
Another popular approach that also satisfies the requirement for adaptivity is Conformalized Quantile Regression (CQR) \cite{romano2019conformalized}. Unlike previous methods that rely on a single point estimate $\hat{Q}_t^{a,i}$, CQR leverages training data $\mathcal{I}_{\mathrm{train}} $ to train two quantile regressors, $\hat{L}(Z_t^{a,i})$ and $\hat{U}(Z_t^{a,i})$, which are trained to estimate the lower and upper quantiles of the target distribution (e.g., at levels $\alpha/2$ and $1-\alpha/2$). For our CNN training step, this means we must replace the RMSE loss function with the pinball loss (also known as quantile loss). The network architecture remains identical to the RMSE based model; only the loss function is modified. In the quantile setting, the output layer is adapted to predict the desired quantile levels, and the model is trained using the pinball loss instead of the mean squared error. The nonconformity score for CQR is defined as 
\begin{equation}\label{eq:scores:CQR}
    S_{\text{CQR}}(Z_t^{a,i}, Q_t^{a,i}) = \max \{ \hat{L}(Z_t^{a,i}) - Q_t^{a,i}, Q_t^{a,i} - \hat{U}(Z_t^{a,i}) \}.
\end{equation}
This score is positive if the true value $Q_t^{a,i}$ falls outside the estimated bounds and negative if it falls inside. The final prediction interval is constructed by shifting the initial quantile estimates:
\begin{equation}\label{eq:intervals:CQR}
    \mathrm{PI}^{\text{CQR}}(Z_t^{a,\text{new}}, \alpha)  = [\hat{L}(Z_t^{a,\text{new}}) - \hat{q}_{1-\alpha} , \hat{U}(Z_t^{a,\text{new}}) + \hat{q}_{1-\alpha} ].
\end{equation}
Here, the conformal quantile appears as a correction term applied to the interval produced by the quantile regressor. In particular, a positive calibration quantile indicates that, for a substantial fraction of calibration points, the true labels fall outside the predicted quantile intervals. Consequently, the intervals must be expanded to achieve the desired coverage.
\subsection{CP Experimental Setup}\label{sec:cp-exp}
In our experiments, we generate evaluation data using 10 random seeds, similar to what is explained in \autoref{sec:Data}. We then randomly select 5 seeds for calibration and the other 5 seeds for testing. We apply CP independently for each of the three variables $(u, h, r)$ and for each assimilation cycle. This leads to $5\times N \times n$ calibration instances per variable–assimilation cycle, and $5\times N \times n$ test instances.
Furthermore, we consider two different variants for applying conformal prediction: ensemble-based and mean-based approaches. In the ensemble-based approach, we consider the entire $5\times N \times n$ instances and apply CP accordingly. In the mean-based approach, we first take the mean across the ensemble size, and then apply CP to the $5\times n$ instances. In either case, we treat the test sets similarly. Also, as rainfall is a non-negative variable, the lower bound of the prediction interval is clipped to zero to ensure positivity of rain.

\section{Inclusion of CP intervals in data assimilation cycling}
\begin{figure}[h!]
\begin{center}
\includegraphics[width=0.45\columnwidth]{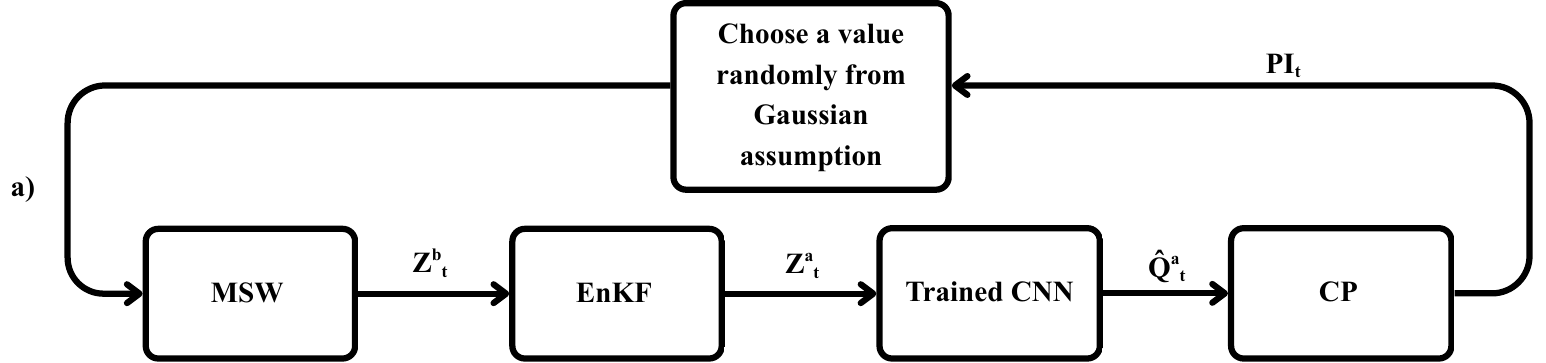}
\includegraphics[width=0.45\columnwidth]{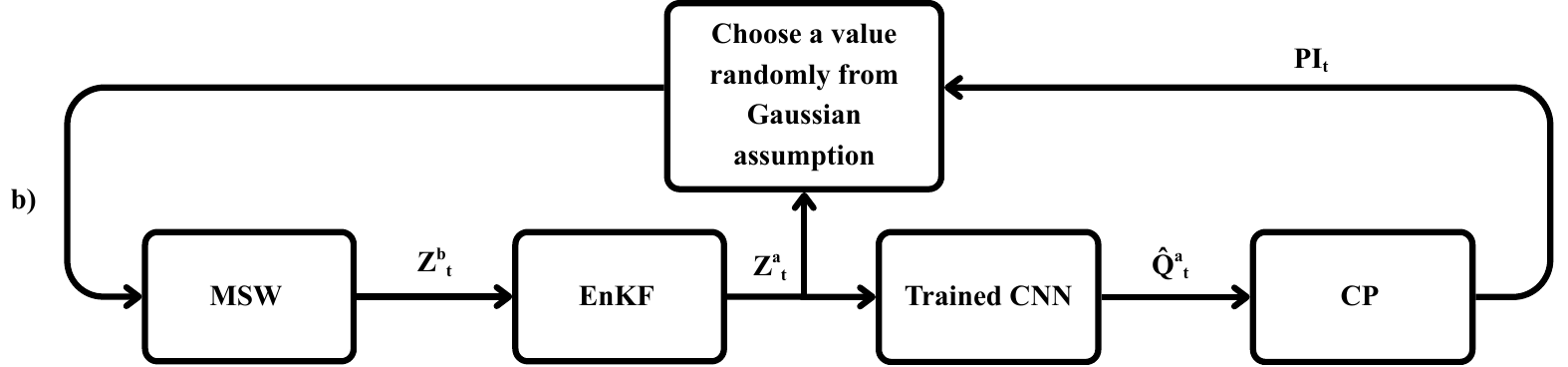}
\caption{Schematic overview of the integration of CP with the CNN-EnKF data assimilation framework, showing perturbation injection to the (a) CNN output and (b) EnKF analysis stages }
\label{CP&DA}
\end{center}
\end{figure}

After training the CNN and obtaining the uncertainty estimates, we investigated the possibility of using these estimates during data assimilation with EnKF. The uncertainty of the hybrid of a CNN and the EnKF has been represented through the analysis ensemble during data assimilation. Therefore, thus far, the CP intervals are constructed in a post-processing way, meaning that the data assimilation cycle operates independently of CP, while CP utilizes the resulting analysis to quantify uncertainty as shown in \autoref{CP}. In particular, CP provides an interval where we expect the QPEns analysis, used here as a proxy truth or reference, to lie for every grid point across all assimilation cycles. Due to the small ensemble size, the ensemble spread alone does not adequately represent the full analysis uncertainty, motivating the use of CP to provide a more reliable uncertainty quantification. Furthermore, since the CNN is trained to approximate QPEns and not the nature run truth, the CP intervals quantify uncertainty with respect to the QPEns proxy truth, and not to the nature run. Therefore, we additionally perturb each analysis ensemble member assuming a Gaussian distribution over the CP interval in each grid point independently, creating the CP informed analysis ensemble 
$\{\tilde{\textbf{z}}^{a,i}\}_{i=1}^N$, i.e.,
\begin{equation}\label{eq:perturbation}
\tilde{\textbf{z}}^{a,i}={\textbf{z}}^{a,i}+{\bf \eta}^i
\end{equation}
Here ${\bf \eta}^i$ is a vector of size of the state, whose each element is a random realization of a normally distributed random variable with mean zero and with standard deviation half of the calculated CP interval for each cycle averaged spatially as mentioned in \autoref{eq:CP_perturbation}.
\begin{equation}\label{eq:CP_perturbation}
    \eta^i = \frac{\mathcal{C}_t}{2}\,\xi^i , 
    \qquad \xi^i\sim \mathcal{N}(0,\,1) ,
\end{equation}
where $\mathcal{C}_t$ is the width of the CP interval at each cycle $t$ and $i = 1,\ldots,N$ indexes the members of the ensemble. Note that for the rainfall variable, the conformal scores are computed without clipping the lower bound of the prediction interval to zero to preserve the full uncertainty in the calibration process. \autoref{CP&DA} provides a schematic overview of the proposed workflow, in which CP intervals are injected as perturbations into the data assimilation cycle. The perturbations are applied either to (a) the CNN analysis or (b) the EnKF analysis, followed by EnKF assimilation. The resulting EnKF analysis is then used by CNN for correcting, which yields updated CNN and EnKF analyzes that incorporate conformal uncertainty information. These two configurations represent different ways of integrating CP within the data assimilation framework, moving beyond the traditional use of CP as a purely post-processing uncertainty quantification method. In addition, investigates how incorporating CP estimates directly within the data assimilation cycle can improve the quality of the analysis, rather than being applied only after the prediction stage.
\section{Results}
We repeat the random splitting of the evaluation seeds into calibration and test, as explained in \autoref{sec:cp-exp}, 10 times, each time performing calibration with the two variants (i.e., ensemble-based and mean-based), for the three algorithms (SCP, NCP, and CQR), for each variable–assimilation cycle, and then evaluating the prediction sets on the test data.
For a given test data point $(Z_t^{a,\text{new}}, Q_t^{a,\text{new}})$ and a prediction interval $\mathrm{PI}(Z_t^{a,\text{new}}, \alpha)$, we compute the coverage as $\mathbb{1}(Q_t^{a,\text{new}} \in \mathrm{PI}(Z_t^{a,\text{new}}, \alpha))$, the interval size as $\max \mathrm{PI}(Z_t^{a,\text{new}}, \alpha) - \min \mathrm{PI}(Z_t^{a,\text{new}}, \alpha)$, miss low which measures the magnitude by which the true value falls below the lower bound of the prediction interval as $\min \mathrm{PI}(Z_t^{a,\text{new}}, \alpha) - Q_t^{a,\text{new}} \cdot \mathbb{1} \{Q_t^{a,\text{new}}< \min \mathrm{PI}(Z_t^{a,\text{new}}, \alpha)\}$, miss high which measures the magnitude by which the true value exceeds the upper bound of the prediction interval as $Q_t^{a,\text{new}} - \max \mathrm{PI}(Z_t^{a,\text{new}}, \alpha) \cdot \mathbb{1} \{Q_t^{a,\text{new}} > \max \mathrm{PI}(Z_t^{a,\text{new}}, \alpha)\}$ and the interval score loss (ISL) \citep{Gneiting2007StrictlyPS} as:
\begin{align}
    \bigg[ &\left(\max \mathrm{PI}(Z_t^{a,\text{new}}, \alpha) - \min \mathrm{PI}(Z_t^{a,\text{new}}, \alpha)\right) \notag\\ &+ \frac{2}{\alpha} \cdot \left(\min \mathrm{PI}(Z_t^{a,\text{new}}, \alpha) - Q_t^{a,\text{new}}\right) \cdot \mathbb{1} \{Q_t^{a,\text{new}}< \min \mathrm{PI}(Z_t^{a,\text{new}}, \alpha)\} \notag\\ &+ \frac{2}{\alpha} \cdot \left(Q_t^{a,\text{new}} - \max \mathrm{PI}(Z_t^{a,\text{new}}, \alpha)\right) \cdot \mathbb{1} \{Q_t^{a,\text{new}} > \max \mathrm{PI}(Z_t^{a,\text{new}}, \alpha)\} \bigg].
\end{align}
For each random split seed, these values are averaged over all test points to calculate the marginal coverage, average interval size, and average interval score loss (AISL) for the target coverage value $1 - \alpha = 0.90$ ($\alpha$ is 0.1 for 90\% coverage). Finally, these average values are aggregated across different random split seeds, and the resulting mean and standard deviation are reported accordingly. In the next subsections 5.1-5.3, different CP methods are compared with respect to average coverage and average interval size, the ensemble based approach is considered first.



\subsection{Uncertainty quantification with SCP}
In this approach, the nonconformity score is computed from the absolute error between the CNN prediction and the QPEns analysis, which is considered the reference solution that approximates the truth. As a result, the CP interval directly reflects how far the CNN is expected to deviate from the QPEns. Larger intervals indicate time steps or spatial regions where the CNN predictions become less certain, whereas narrow intervals correspond to high local agreement with QPEns.
\begin{figure}[h!]
\begin{center}
\includegraphics[width=0.70\columnwidth]{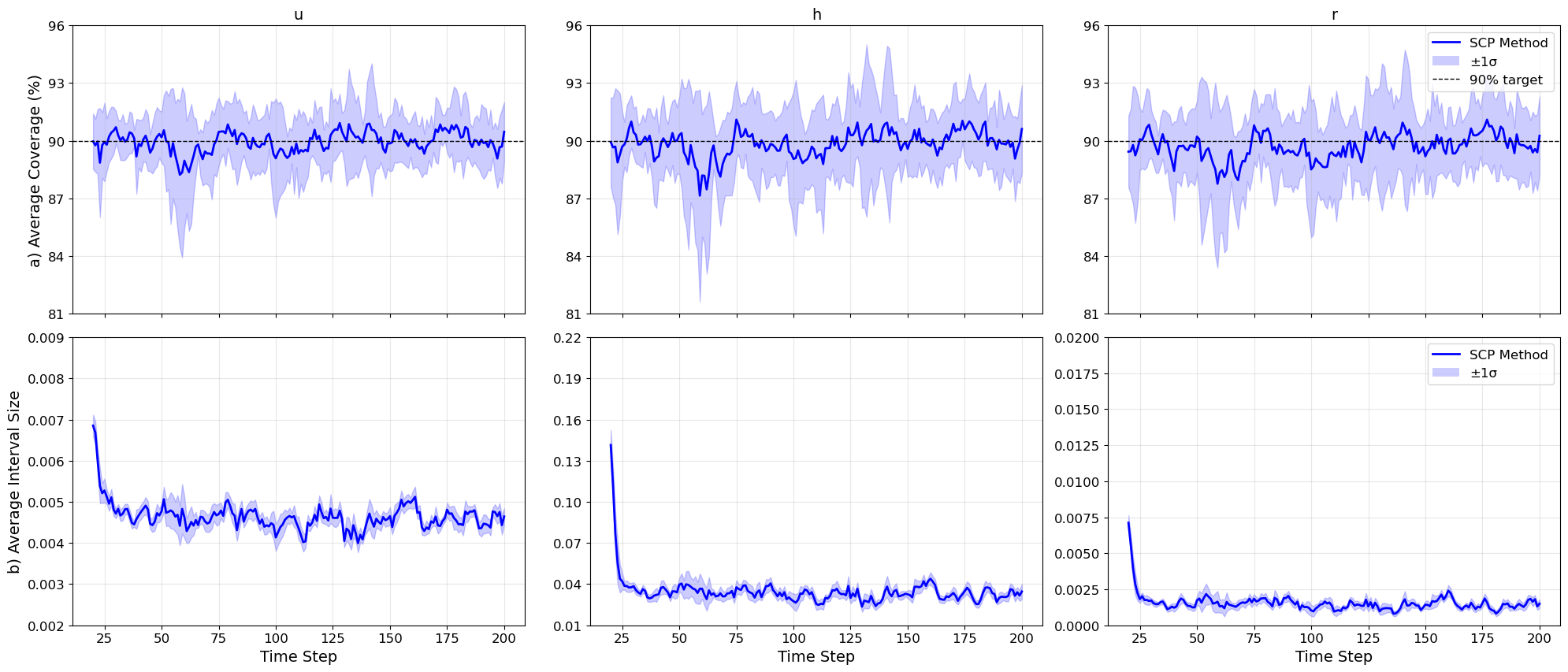}
\caption{Temporal evolution of the a) average empirical coverage and b) average interval size for SCP method applied to variables $u$ (left), $h$ (middle), $r$ (right) from timestep 20. In a) solid line denote mean ensemble coverage across random seeds, shaded region indicate coverage variability, and the dashed black line represents the nominal 90\% target coverage. In b) solid line denote mean ensemble interval size across random seeds and the shaded region indicate interval size variability. }
\label{SCP_ci}
\end{center}
\end{figure}

Figure~\ref{SCP_ci}a, shows the average empirical coverage over all assimilation steps for the three prognostic variables u, h, r. Here, average empirical coverage denotes the fraction of grid points whose reference states are contained within the CP intervals, averaged over all ensemble members and random seeds at each time step. Across the variables, the ensemble based SCP consistently achieves coverage close to the nominal 90\% level, demonstrating that the calibrated nonconformity distribution yields well-balanced uncertainty intervals. Coverage for the variables u, h, r remains particularly stable but display slightly higher variability around timestep 50 to 75. Nevertheless, no systematic over or under coverage occurs, indicating that SCP provides a robust baseline uncertainty representation across the full assimilation window. The average SCP interval sizes (Figure~\ref{SCP_ci}b) reveal how uncertainty evolves over time. The initial jump in the interval size is due to the spin-up time for the data assimilation methods to stabilize, which can be ignored. For u, the interval sizes remain very small (between $4 \times 10^{-3}$  and $5.5 \times 10^{-3}$ ), indicating increased deviation and less smooth velocity field. The height field (h) displays intervals around $2.5 \times 10^{-2}$  and $5 \times 10^{-2}$. The rainfall variable r widths up to roughly $1.5 \times 10^{-3}$  and $2.5 \times 10^{-3}$ .
\begin{figure}[h!]
\begin{center}
\includegraphics[width=0.45\columnwidth]{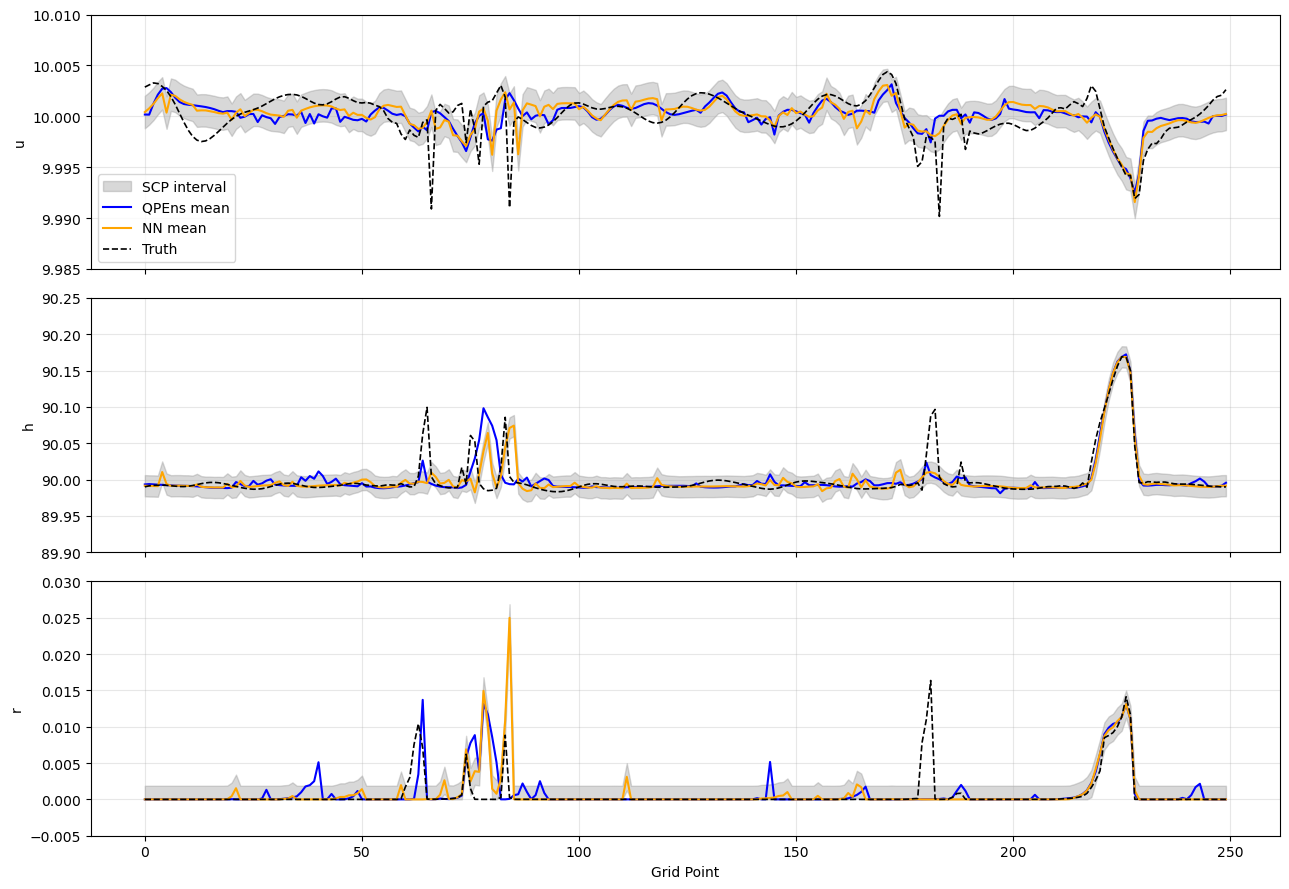}
\includegraphics[width=0.45\columnwidth]{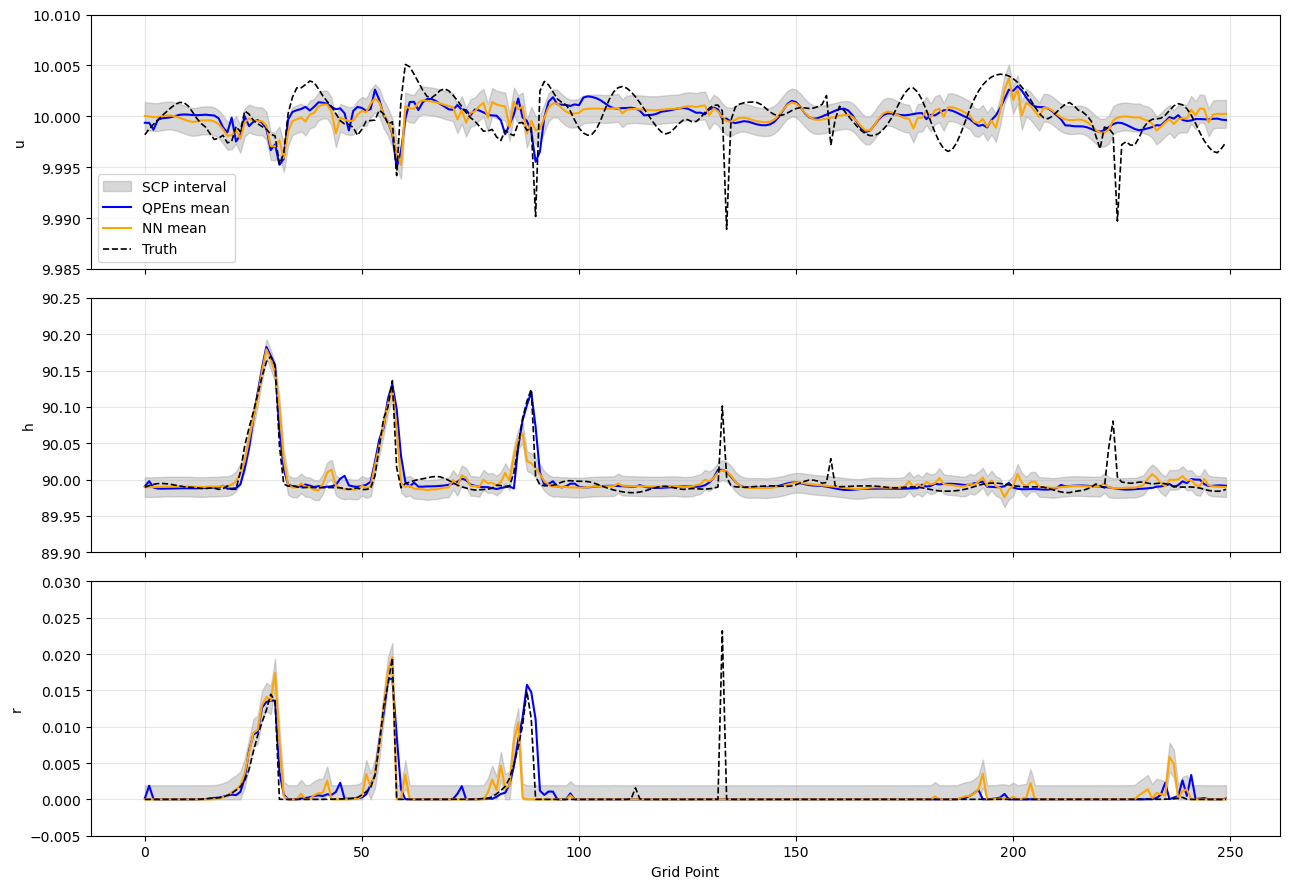}
\caption{Truth (black dashed) and ensemble mean for QPEns (blue) and NN (orange) at timestep 25 (left) and 160 (right). Gray shading represents uncertainty calculated with SCP Method for the three prognostic variables $u$ (upper), $h$ (middle), $r$ (lower).}
\label{SCP_25}
\end{center}
\end{figure}

To illustrate how the SCP intervals influence the physical representation of each state variable, we present spatial snapshots at two representative time steps. The early snapshot (t=25) captures the initial assimilation phase, while the later snapshot (t=160) reflects long-term behaviour after the system has adjusted to repeated assimilation. Together, these snapshots provide a clear visualization of the evolution of uncertainty and physical consistency between variables. The snapshot at timesteps 25 and 160 (Figure~\ref{SCP_25}) shows the ensemble mean of QPEns and CNN as well as the true fields together with the SCP intervals for a randomly selected test sample. For variable r, the interval width is clipped to zero to showcase the positivity of rain. The CP interval for each variable remains essentially constant across all grid points because SCP uses a single global nonconformity threshold and does not adapt locally to variations in spatial scale. As a result, while the interval width is appropriate for most regions, at a few grid points typically near sharper peaks, the interval does not fully enclose the QPEns value, and the truth also lies outside the bounds. This behaviour of truth field not captured fully is consistent with the computation of SCP, which calibrates uncertainty solely based on CNN-QPEns deviations, and also this method cannot adjust its width to local dynamical variability. These limitations motivate the use of other CP methods, which rescales the nonconformity score and allows the interval width to vary across space.

\subsection{Uncertainty quantification with NCP}
The NCP results show how incorporating a scaling term enables the interval widths to adapt locally across both time and space, producing uncertainty bands that better reflect the structure of each variable. Here, the nonconformity score is scaled by a normalization factor ‘$\sigma$’, defined as the standard deviation of the CNN ensemble predictions across grid points. The normalization factor $\sigma$ is estimated using a Random Forest regressor composed of 100 decision trees with a maximum depth of 20. The model is trained separately for each variable to predict the magnitude of the absolute residual between the CNN and QPEns analyses, with five independent random seeds to improve robustness. The resulting $\sigma$ values are averaged across seeds and used to normalize the residuals in the computation of nonconformity scores, allowing the method to account for heteroscedastic uncertainty. NCP is applied only to the variables u and h, as the rainfall variable r contains many zero valued grid points; using normalization in such regions would lead to unstable or inflated nonconformity scores.  
\begin{figure}[h!]
\begin{center}
\includegraphics[width=0.70\columnwidth]{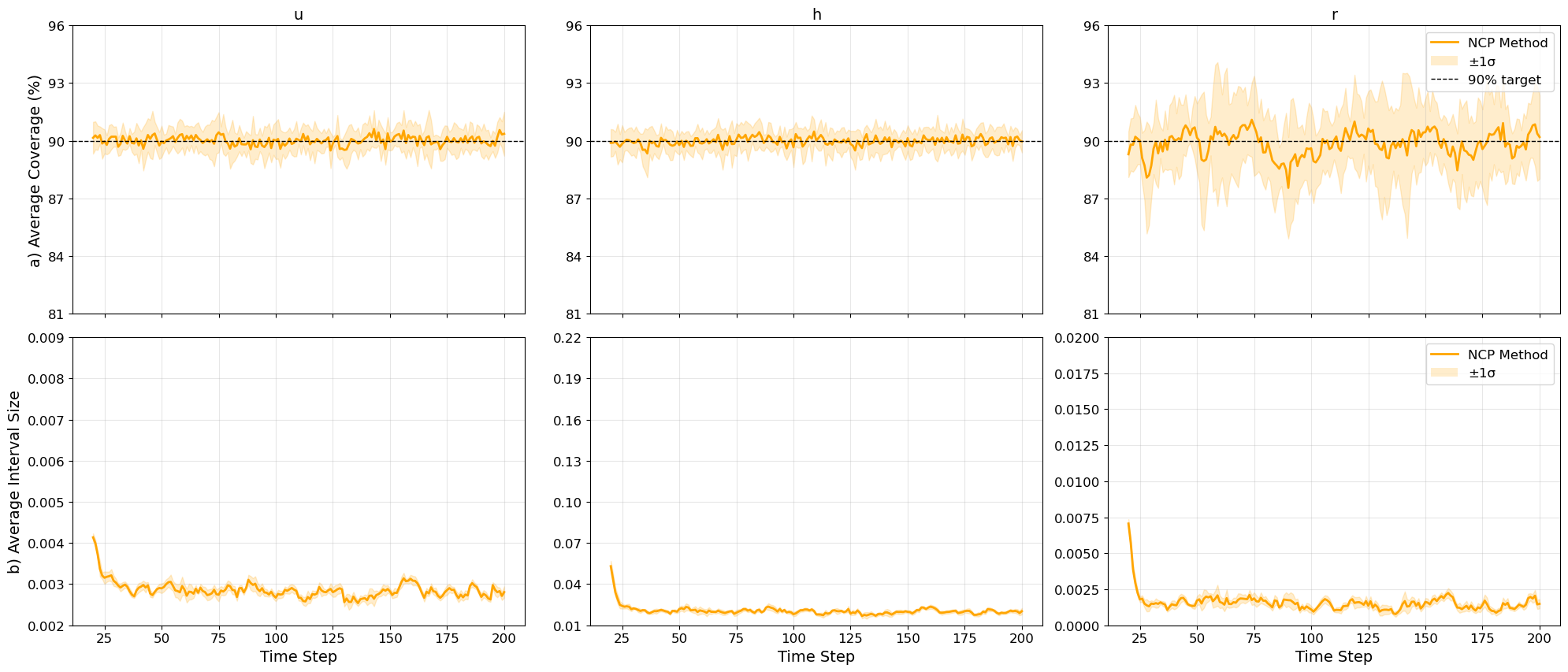}
\caption{Temporal evolution of the a) average empirical coverage and b) average interval size for NCP method applied to variables $u$ (left), $h$ (middle), $r$ (right) from timestep 20. In a) solid line denote mean ensemble coverage across random seeds, shaded region indicate coverage variability, and the dashed black line represents the nominal 90\% target coverage. In b) solid line denote mean ensemble interval size across random seeds and the shaded region indicate interval size variability.}
\label{NCP_ci}
\end{center}
\end{figure}

\begin{figure}[h!]
\begin{center}
\includegraphics[width=0.45\columnwidth]{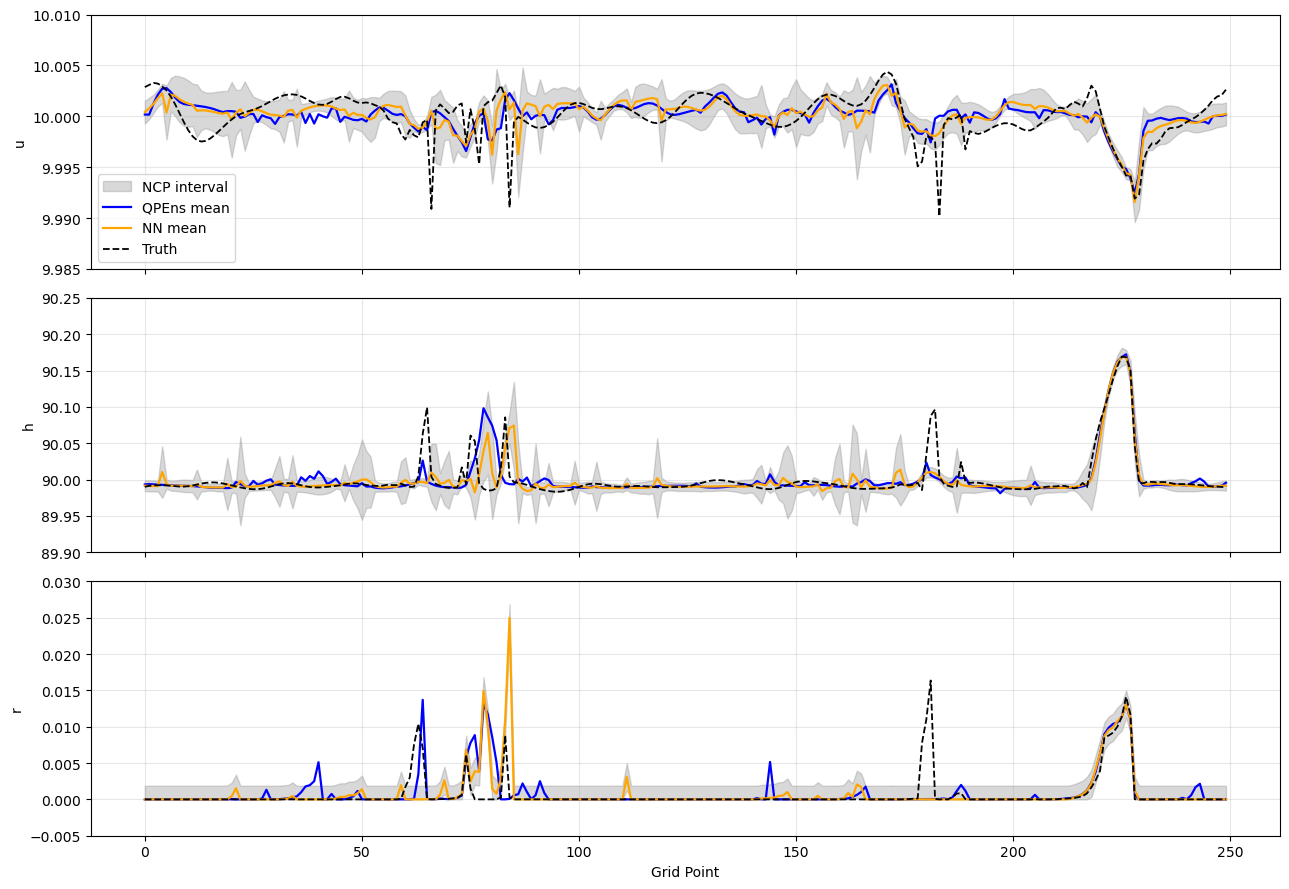}
\includegraphics[width=0.45\columnwidth]{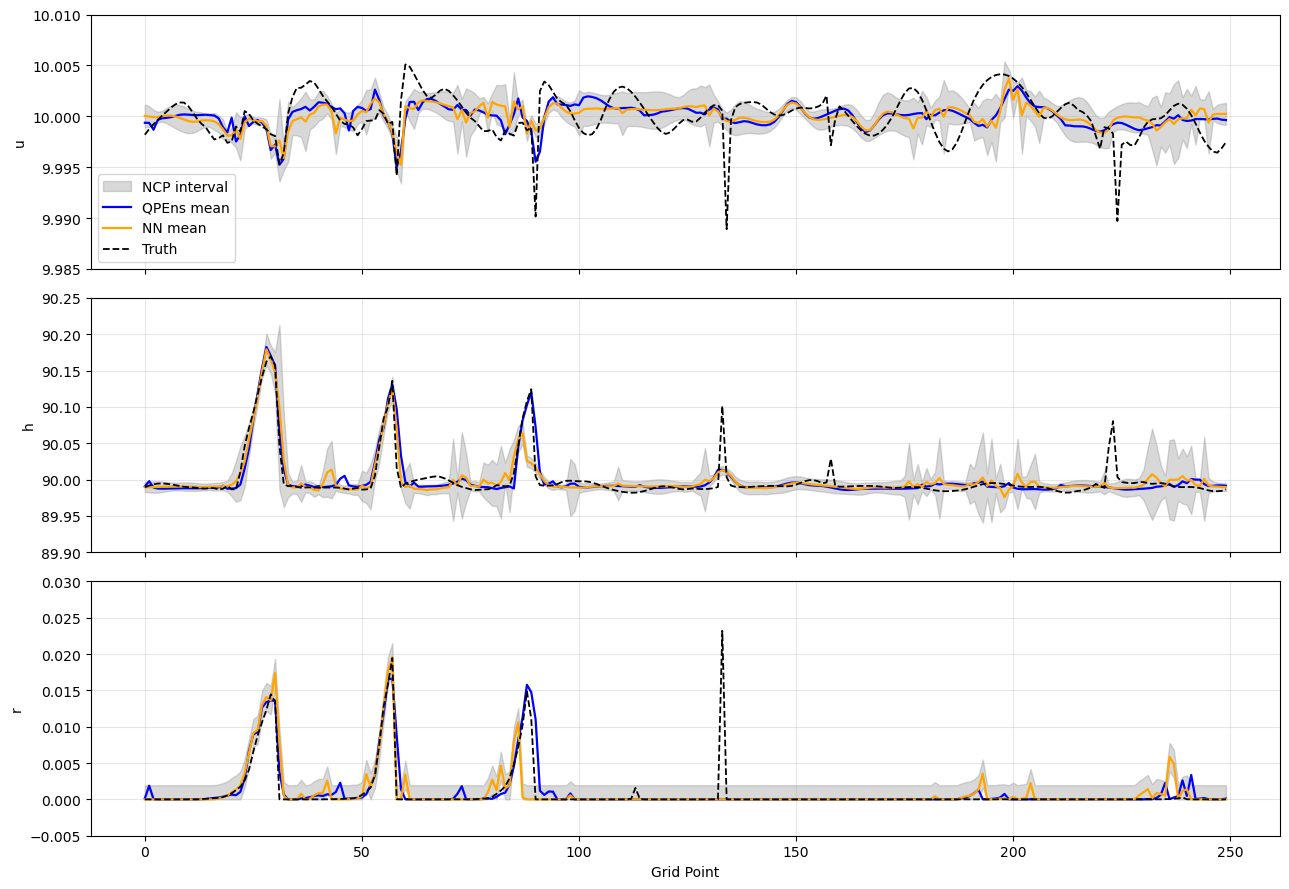}
\caption{Truth (black dashed) and ensemble mean for QPEns (blue) and NN (orange) at timestep 25 (left) and 160 (right). Gray shading represents uncertainty calculated with NCP Method for the three prognostic variables $u$ (upper), $h$ (middle), $r$ (lower).}
\label{NCP_25}
\end{center}
\end{figure}

In Figure~\ref{NCP_ci}a, the ensemble based NCP maintains average empirical coverage close to the nominal 90\% level for all three variables, with noticeably varying intervals compared to the SCP. The average interval sizes (Figure~\ref{NCP_ci}b) for the velocity component u is around $2.4 \times 10^{-3}$  and $3.2 \times 10^{-3}$, the variable h around $1.5 \times 10^{-2}$  and $2.2 \times 10^{-2}$ , and the rainfall component near $1.5 \times 10^{-3}$ and $2.5 \times 10^{-3}$ . These values are smaller than those produced by the SCP because the normalization term compensates for local changes in space.
The snapshot of the ensemble mean of a randomly selected test sample at timestep 25 and 160 (Figure~\ref{NCP_25}) further illustrates this behaviour; unlike SCP which imposes a uniform interval width across all grid points, NCP expands the bands in regions with stronger local variability and contracts them in smoother areas, producing spatially responsive uncertainty estimates; although a few sharp peaks remain outside the interval for certain grid points. 

\subsection{Uncertainty quantification with CQR}
\begin{figure}[h!]
\begin{center}
\includegraphics[width=0.70\columnwidth]{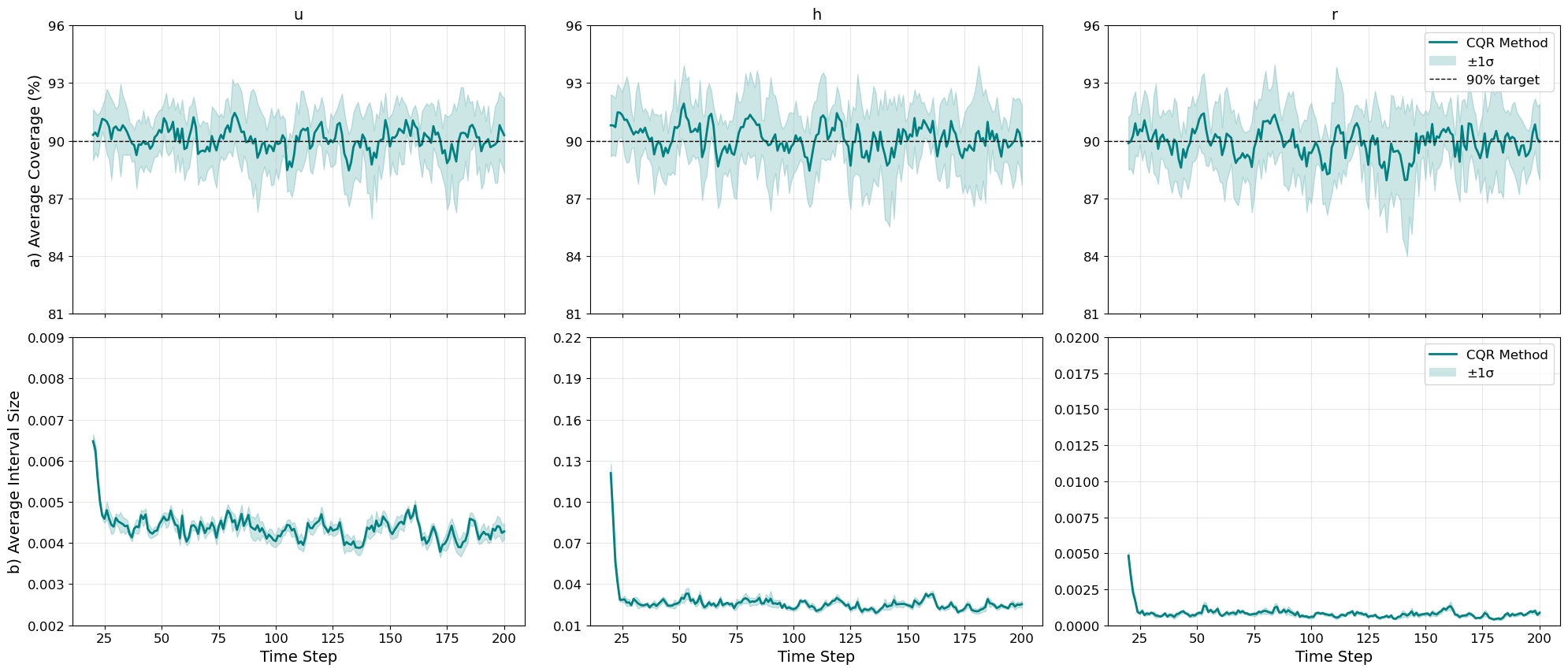}
\caption{Temporal evolution of the a) average empirical coverage and b) average interval size for CQR method applied to variables $u$ (left), $h$ (middle), $r$ (right) from timestep 20. In a) solid line denote mean ensemble coverage across random seeds, shaded region indicate coverage variability, and the dashed black line represents the nominal 90\% target coverage. In b) solid line denote mean ensemble interval size across random seeds and the shaded region indicate interval size variability.}
\label{CQR_ci}
\end{center}
\end{figure}

The CQR framework is constructed by training a quantile regression neural network using the pinball loss at two quantile levels, $1-\alpha/2$= 0.95 and $\alpha/2$=0.05, enabling the model to learn data driven lower and upper predictive bounds. In Figure~\ref{CQR_ci}a, the ensemble based CQR method maintains average coverage close to the prescribed 90\% level for all variables but more variability than NCP for variables u and h. The average interval sizes (Figure~\ref{CQR_ci}b) follow a pattern like the earlier CP variants: an initial rise during spin-up followed by a rapid contraction after timestep 20. The stabilized widths lie around $3.8 \times 10^{-3}$ and $5 \times 10^{-3}$  for u, approximately $2 \times 10^{-2}$ to $3.5 \times 10^{-2}$  for h, and roughly $1 \times 10^{-3}$ and $2 \times 10^{-3}$  for r. 
\begin{figure}[h!]
\begin{center}
\includegraphics[width=0.45\columnwidth]{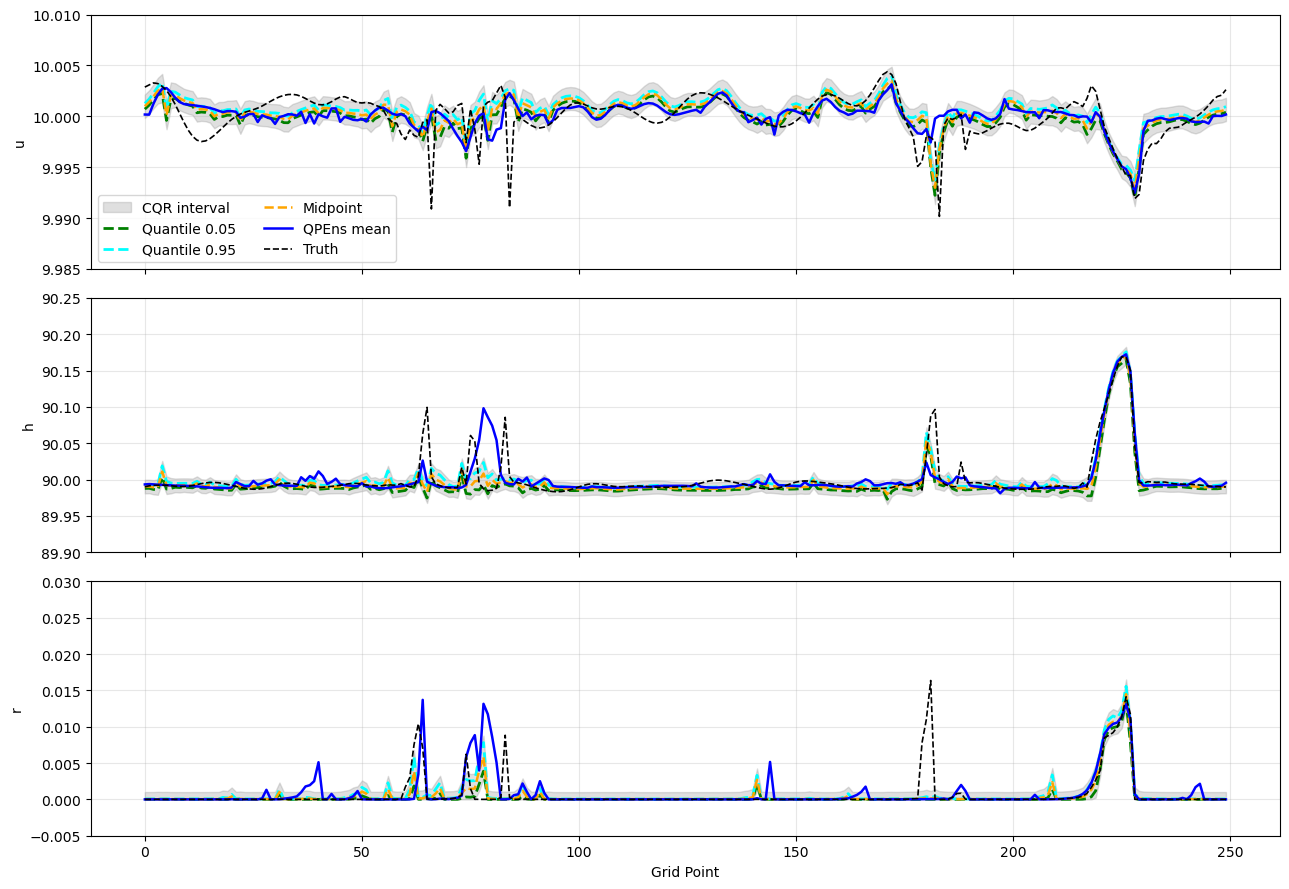}
\includegraphics[width=0.45\columnwidth]{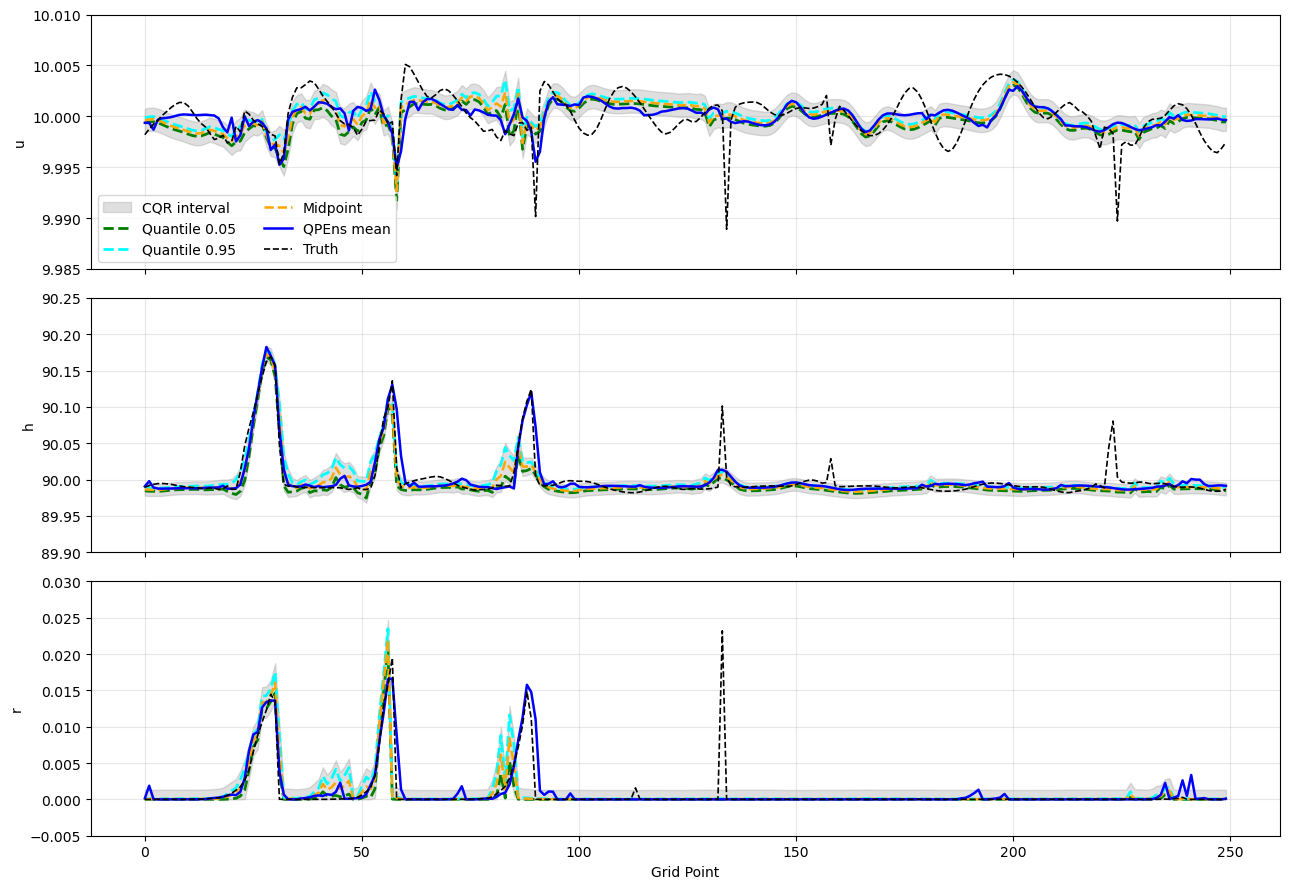}
\caption{Truth (black dashed) and ensemble mean of QPEns (blue),upper quantile ($1-\alpha/2$=0.95,cyan), lower quantile ($\alpha/2$=0.05,green) and midpoint of quantiles (orange) at timestep 25 (left) and 160 (right). Gray shading represents uncertainty calculated with CQR Method for the three prognostic variables $u$ (upper), $h$ (middle), $r$ (lower).}
\label{CQR_25}
\end{center}
\end{figure}
The spatial snapshot at timestep 25 and 160 (Figure~\ref{CQR_25}) illustrates how CQR forms locally responsive intervals. The quantile bands widen near regions where the QPEns solution exhibits stronger deviations from the median, while contracting in smoother segments of the domain. The quantile outputs appear in the figure as dashed green ($\alpha/2$=0.05) and dashed cyan ($1-\alpha/2$=0.95) lines, with their midpoint shown in green. For the rainfall variable r, the behaviour is notably different when compared to other methods: because r is highly sporadic and contains many zero valued grid points, the network learns a very narrow lower quantile and a rapidly expanding upper quantile near convective peaks, resulting in visibly asymmetric intervals. This reflects the innate skewness and heavy tailed distribution of the rainfall field rather than instability in the CQR method. The interval for u is noticeably wider than for h, reflecting larger spread in the velocity component, while for h, the quantiles stay relatively close and only widen slightly around the regions of increased variability.

\subsection{Comparison of uncertainty quantification in different methods }
 In this section, the performance of different uncertainty quantification approaches is analyzed through two complementary comparisons as mentioned in \autoref{sec:cp-exp}. First, the behaviour of the three CP methods is examined by computing CP intervals using two different strategies: computing CP directly from the ensemble members and computing CP from the ensemble mean. This allows us to evaluate how the method of constructing the CP scores influences the resulting uncertainty estimates and to assess which formulation provides more reliable results. In the second comparison, CP-based uncertainty estimates computed from the ensemble mean are evaluated against traditional uncertainty quantification methods, namely the standard deviation interval and the ensemble mean spread. This comparison provides insight into how CP methods perform relative to conventional approaches and illustrates how CP can be incorporated as an alternative way to quantify predictive uncertainty.

\subsubsection{Comparison of uncertainty quantification in CP methods }
From Table~\ref{Aisl_QPEns} which summarize the AISL values, empirical coverage percentages, interval width and directional error rates (miss-high and miss-low) averaged over the full assimilation window computed against QPEns, the average score loss interval (AISL) matrix \cite{Gneiting2007StrictlyPS} provides a quantitative view of how each CP method behaves across the three prognostic variables. Here, we are focusing on comparing the three CP methods when the conformal scores are computed directly from the ensemble members and from the ensemble mean. Both formulations provide reliable uncertainty estimates with coverage generally close to the nominal 90\% target. The comparison is therefore intended not to demonstrate failure of either approach, but rather to identify which formulation offers a more stable and representative implementation of CP-based uncertainty quantification within the ensemble forecasting framework.
For SCP, the ensemble-mean formulation generally provides improved uncertainty estimates compared to the ensemble-based implementation. The SCP mean-based approach consistently produces smaller AISL values, such as 0.004$\pm$0.001 for u compared to 0.008$\pm$0.001 for the ensemble-based formulation, while also maintaining coverage closer to the nominal target (89.93\%$\pm$4.18\% compared to 89.89\%$\pm$3.02\%). The interval widths are also slightly narrower for the mean-based approach (0.002\% for u) than for the ensemble-based formulation (0.004\%). A similar behaviour is observed for h, where AISL decreases from 0.094$\pm$0.027 to 0.055$\pm$0.023. For variable r, AISL is lesser for mean based (0.008) than ensemble based (0.012), but interval width is slightly higher by 0.001 with near similar miss rates. Overall, the SCP mean-based formulation provides slightly tighter intervals while maintaining balanced uncertainty estimates.
For NCP, the opposite trend is observed, with the ensemble-based NCP generally providing stronger performance than its mean-based counterpart. For u, the ensemble-based implementation achieves an AISL of 0.003$\pm$0.000 with coverage of 89.98\%$\pm$1.36\%, compared to 0.004$\pm$0.001 and 90.02\%$\pm$3.96\% for the mean-based approach. A similar pattern is seen for h, where the ensemble-based method produces AISL of 0.021$\pm$0.005 compared to 0.052$\pm$0.019 for the mean-based method, while maintaining coverage close to the nominal level. The interval widths are also same for both the implementations, and the miss-low and miss-high rates remain relatively symmetric for ensemble based approach and highly asymmetric for mean based approach. But for the variable r, the AISL value for mean based is lower (0.008) but has higher interval width by 0.001 when compared with the ensemble based counterpart.
For CQR, the mean-based formulation tends to provide improved performance compared to the ensemble-based version. For instance, for u, the AISL decreases from 0.007$\pm$0.001 in the ensemble-based formulation to 0.004$\pm$0.001 when using the ensemble mean, while maintaining coverage close to 90\%. A similar improvement is observed for h, where the AISL decreases from 0.080$\pm$0.026 to 0.051$\pm$0.019. For the variable r, mean based have lower AISL values of 0.007 than ensemble based with 0.011 and similar coverage levels. However, the mean-based CQR approach generally produces slightly narrower intervals. 

To further assess the robustness of the CP-based uncertainty estimates, the same metrics are also evaluated against the true model state rather than QPEns (Table~\ref{Aisl_Truth}). Since the CP intervals are constructed using QPEns as the reference, the empirical coverage with respect to truth is not expected to match the nominal 90\% level. Instead, these statistics provide insight into how well the QPEns-based uncertainty representation captures the actual model error.
For SCP, the mean-based formulation again shows more consistent performance when evaluated against truth. For u, the mean formulation achieves AISL of 0.010$\pm$0.002 but with lower coverage of 64.46\%, while the ensemble-based implementation produces a slightly larger AISL (0.011$\pm$0.001) and lower coverage (77.99\%). The directional miss rates also portray this difference in the ensemble formulation (11.44\% miss-low and 10.51\% miss-high) compared to the mean-based case (17.99\% and 17.55\%). For the variable h, similarly mean based have lower AISL values of 0.073 than ensemble based with 0.094, with ensemble-based approach maintaining higher interval width (0.034) than the mean-based formulation (0.024). For the variable r, both approaches produce comparable coverage levels but lower AISL for mean based with slightly higher interval width than ensemble based.
For NCP, evaluating against truth reveals larger deviations, particularly for the ensemble-based implementation. For u, the ensemble formulation yields AISL of 0.016$\pm$0.001 but only 44.28\% coverage, accompanied by large directional miss rates (29.82\% miss-low and 25.90\% miss-high). The mean-based implementation improves the AISL (0.010$\pm$0.001) and increases coverage to 62.83\%, although the miss rates remain relatively high. A similar improvement is observed for the variable h, where the ensemble-based formulation achieves 47.41\% coverage compared to 80.35\% for the mean-based version. For the variable r, however, both implementations perform similarly, with AISL values lower for mean based (0.009), but both approaches with coverage close to 90\% and interval width around 0.001-0.002.
For CQR, the both implementations demonstrate near similar performances. For u, both formulation produces AISL of 0.011$\pm$ but ensemble based have coverage of 75.18\% coverage, while the mean-based formulation yields lower coverage of 61.70\% despite closer AISL values. For h, mean based have lower AISL of 0.068 than ensemble based with 0.078, though the ensemble formulation maintains slightly higher coverage (85.12\%) than the mean-based version (84.70\%). For the rainfall r, both implementations perform well, with AISL values around 0.009 and coverage around 90\%, while maintaining relatively small directional miss rates.

\begin{table}[h!]
\centering
\small
\rowcolors{0}{}{}
\resizebox{\textwidth}{!}{
\begin{tabular}{l c c c c c c}
\rowcolor{gray!20}
Method & Variable & AISL (mean$\pm$std) & Coverage (mean$\pm$std) & Interval Width (mean$\pm$std) & Miss Low (mean$\pm$std) & Miss High (mean$\pm$std)\\
\hline
SCP ensemble based& u & 0.008 $\pm$ 0.001 & 89.89\% $\pm$ 3.02\% & 0.004 $\pm$ 0.000 & 4.94\% $\pm$ 1.54\% & 5.17\% $\pm$ 1.84\% \\
 & h & 0.094 $\pm$ 0.027 & 89.82\% $\pm$ 3.88\% & 0.034 $\pm$ 0.012 & 5.22\% $\pm$ 2.31\% & 4.95\% $\pm$ 1.96\% \\
 & r & 0.012 $\pm$ 0.005 & 89.64\% $\pm$ 3.85\% & 0.001 $\pm$ 0.000 & 5.71\% $\pm$ 2.39\% & 4.65\% $\pm$ 1.93\% \\
NCP ensemble based& u & 0.003 $\pm$ 0.000 & 89.98\% $\pm$ 1.36\% & 0.002 $\pm$ 0.000 & 5.19\% $\pm$ 0.94\% & 4.83\% $\pm$ 0.97\% \\
 & h & 0.021 $\pm$ 0.005 & 89.98\% $\pm$ 1.14\% & 0.020 $\pm$ 0.005 & 4.73\% $\pm$ 0.93\% & 5.29\% $\pm$ 1.01\% \\
 & r & 0.012 $\pm$ 0.005 & 89.64\% $\pm$ 3.85\% & 0.001 $\pm$ 0.000 & 5.71\% $\pm$ 2.39\% & 4.65\% $\pm$ 1.93\% \\
CQR ensemble based& u & 0.007 $\pm$ 0.001 & 89.87\% $\pm$ 2.99\% & 0.004 $\pm$ 0.000 & 5.84\% $\pm$ 1.77\% & 4.29\% $\pm$ 1.67\% \\
 & h & 0.080 $\pm$ 0.026 & 89.84\% $\pm$ 3.62\% & 0.026 $\pm$ 0.009 & 3.49\% $\pm$ 2.09\% & 6.66\% $\pm$ 2.27\% \\
 & r & 0.011 $\pm$ 0.000 & 89.84\% $\pm$ 3.63\% & 0.001 $\pm$ 0.000 & 3.02\% $\pm$ 1.60\% & 7.14\% $\pm$ 2.76\% \\
 \\
SCP mean based& u & 0.004 $\pm$ 0.001 & 89.93\% $\pm$ 4.18\% & 0.002 $\pm$ 0.000 & 4.73\% $\pm$ 2.33\% & 5.34\% $\pm$ 2.81\% \\
 & h & 0.055 $\pm$ 0.023 & 89.93\% $\pm$ 4.32\% & 0.024 $\pm$ 0.007 & 5.36\% $\pm$ 2.87\% & 4.71\% $\pm$ 2.35\% \\
 & r & 0.008 $\pm$ 0.004 & 89.85\% $\pm$ 4.20\% & 0.002 $\pm$ 0.000 & 5.49\% $\pm$ 2.76\% & 4.66\% $\pm$ 2.18\% \\
NCP mean based& u & 0.004 $\pm$ 0.001 & 90.02\% $\pm$ 3.96\% & 0.002 $\pm$ 0.000 & 6.11\% $\pm$ 2.60\% & 3.87\% $\pm$ 2.35\% \\
 & h & 0.052 $\pm$ 0.019 & 89.88\% $\pm$ 4.01\% & 0.027 $\pm$ 0.007 & 2.06\% $\pm$ 1.56\% & 8.06\% $\pm$ 3.25\% \\
 & r & 0.008 $\pm$ 0.004 & 89.85\% $\pm$ 4.20\% & 0.002 $\pm$ 0.000 & 5.49\% $\pm$ 2.76\% & 4.66\% $\pm$ 2.18\% \\
CQR mean based& u & 0.004 $\pm$ 0.001 & 89.85\% $\pm$ 4.26\% & 0.002 $\pm$ 0.000 & 6.50\% $\pm$ 3.14\% & 3.64\% $\pm$ 2.26\% \\
 & h & 0.051 $\pm$ 0.021 & 89.90\% $\pm$ 3.92\% & 0.020 $\pm$ 0.006 & 2.20\% $\pm$ 2.01\% & 7.89\% $\pm$ 3.53\% \\
 & r & 0.008 $\pm$ 0.003 & 89.81\% $\pm$ 4.21\% & 0.002 $\pm$ 0.000 & 2.56\% $\pm$ 1.80\% & 7.63\% $\pm$ 3.34\% \\
\\
Std Deviation & u & 0.005 $\pm$ 0.000 & 95.71\% $\pm$ 2.15\% & 0.004 $\pm$ 0.000 & 2.64\% $\pm$ 1.45\% & 1.65\% $\pm$ 1.22\% \\
 & h & 0.051 $\pm$ 0.019 & 88.80\% $\pm$ 3.84\% & 0.025 $\pm$ 0.005 & 2.33\% $\pm$ 1.66\% & 8.88\% $\pm$ 3.14\% \\
 & r & 0.007 $\pm$ 0.003 & 31.79\% $\pm$ 8.07\% & 0.001 $\pm$ 0.000 & 1.83\% $\pm$ 1.34\% & 66.38\% $\pm$ 8.48\% \\
\\
Ensemble spread & u & - & 94.53\% $\pm$ 2.48\% & 0.004 $\pm$ 0.000 & 3.12\% $\pm$ 1.58\% & 2.33\% $\pm$ 1.53\% \\
 & h & - & 87.13\% $\pm$ 4.19\% & 0.024 $\pm$ 0.005 & 3.74\% $\pm$ 2.33\% & 9.13\% $\pm$ 3.17\% \\
 & r & - & 31.83\% $\pm$ 7.99\% & 0.002 $\pm$ 0.000 & 2.55\% $\pm$ 1.66\% & 65.62\% $\pm$ 8.62\% \\
\\
\end{tabular}}
\caption{Summary of AISL, empirical coverage percentage, interval width and directional error rates percentage (miss-high and miss-low) for the three CP methods (both considering all ensembles and mean based), standard deviation method and ensemble mean spread method averaged over the full assimilation window computed against QPEns.}
\label{Aisl_QPEns}
\end{table}

\begin{table}[h!]
\centering
\small
\rowcolors{0}{}{}
\resizebox{\textwidth}{!}{
\begin{tabular}{l c c c c c c}
\rowcolor{gray!20}
Method & Variable & AISL (mean$\pm$std) & Coverage (mean$\pm$std) & Interval Width (mean$\pm$std) & Miss Low (mean$\pm$std) & Miss High (mean$\pm$std)\\
\hline
SCP ensemble based& u & 0.011 $\pm$ 0.001 & 78.05\% $\pm$ 4.16\% & 0.004 $\pm$ 0.000 & 11.44\% $\pm$ 2.72\% & 10.51\% $\pm$ 2.46\% \\
 & h & 0.094 $\pm$ 0.026 & 89.50\% $\pm$ 3.65\% & 0.034 $\pm$ 0.012 & 5.60\% $\pm$ 2.08\% & 4.89\% $\pm$ 2.04\% \\
 & r & 0.012 $\pm$ 0.004 & 89.85\% $\pm$ 3.26\% & 0.001 $\pm$ 0.000 & 6.67\% $\pm$ 2.18\% & 3.48\% $\pm$ 1.75\% \\
NCP ensemble based& u & 0.016 $\pm$ 0.001 & 44.28\% $\pm$ 4.01\% & 0.002 $\pm$ 0.000 & 29.82\% $\pm$ 3.33\% & 25.90\% $\pm$ 3.39\% \\
 & h & 0.091 $\pm$ 0.014 & 47.41\% $\pm$ 4.69\% & 0.020 $\pm$ 0.005 & 23.37\% $\pm$ 4.37\% & 29.22\% $\pm$ 4.04\% \\
 & r & 0.012 $\pm$ 0.004 & 89.85\% $\pm$ 3.26\% & 0.001 $\pm$ 0.000 & 6.67\% $\pm$ 2.18\% & 3.48\% $\pm$ 1.75\% \\
CQR ensemble based& u & 0.011 $\pm$ 0.001 & 75.18\% $\pm$ 4.32\% & 0.004 $\pm$ 0.000 & 14.08\% $\pm$ 2.99\% & 10.74\% $\pm$ 2.64\% \\
 & h & 0.078 $\pm$ 0.024 & 85.12\% $\pm$ 4.97\% & 0.026 $\pm$ 0.009 & 4.49\% $\pm$ 2.14\% & 10.39\% $\pm$ 3.86\% \\
 & r & 0.009 $\pm$ 0.004 & 90.59\% $\pm$ 3.00\% & 0.001 $\pm$ 0.000 & 3.94\% $\pm$ 1.49\% & 5.47\% $\pm$ 2.34\% \\
 \\
SCP mean based& u & 0.010 $\pm$ 0.002 & 64.46\% $\pm$ 6.91\% & 0.002 $\pm$ 0.000 & 17.99\% $\pm$ 4.54\% & 17.55\% $\pm$ 4.42\% \\
 & h & 0.073 $\pm$ 0.024 & 86.32\% $\pm$ 4.75\% & 0.024 $\pm$ 0.007 & 7.99\% $\pm$ 3.26\% & 5.69\% $\pm$ 2.23\% \\
 & r & 0.009 $\pm$ 0.004 & 89.55\% $\pm$ 3.82\% & 0.002 $\pm$ 0.000 & 7.76\% $\pm$ 3.05\% & 2.69\% $\pm$ 1.53\% \\
NCP mean based& u & 0.010 $\pm$ 0.001 & 62.83\% $\pm$ 6.94\% & 0.002 $\pm$ 0.000 & 20.18\% $\pm$ 4.47\% & 17.00\% $\pm$ 4.53\% \\
 & h & 0.068 $\pm$ 0.020 & 80.35\% $\pm$ 6.33\% & 0.027 $\pm$ 0.007 & 7.22\% $\pm$ 3.81\% & 12.44\% $\pm$ 4.19\% \\
 & r & 0.009 $\pm$ 0.004 & 89.55\% $\pm$ 3.82\% & 0.002 $\pm$ 0.000 & 7.76\% $\pm$ 3.05\% & 2.69\% $\pm$ 1.53\% \\
CQR mean based& u & 0.011 $\pm$ 0.002 & 61.70\% $\pm$ 6.98\% & 0.002 $\pm$ 0.000 & 21.49\% $\pm$ 4.92\% & 16.81\% $\pm$ 4.57\% \\
 & h & 0.066 $\pm$ 0.022 & 84.70\% $\pm$ 5.44\% & 0.020 $\pm$ 0.006 & 4.51\% $\pm$ 2.88\% & 10.79\% $\pm$ 4.27\% \\
 & r & 0.009 $\pm$ 0.003 & 92.58\% $\pm$ 2.91\% & 0.002 $\pm$ 0.000 & 3.97\% $\pm$ 2.03\% & 3.45\% $\pm$ 1.75\% \\
\\
Std Deviation & u & 0.008 $\pm$ 0.001 & 80.46\% $\pm$ 4.64\% & 0.004 $\pm$ 0.000 & 11.17\% $\pm$ 3.19\% & 8.38\% $\pm$ 3.03\% \\
 & h & 0.068 $\pm$ 0.020 & 77.53\% $\pm$ 5.21\% & 0.025 $\pm$ 0.005 & 8.48\% $\pm$ 3.65\% & 13.99\% $\pm$ 3.79\% \\
 & r & 0.007 $\pm$ 0.003 & 32.42\% $\pm$ 8.78\% & 0.001 $\pm$ 0.000 & 2.56\% $\pm$ 1.22\% & 65.02\% $\pm$ 8.97\% \\
\\
Ensemble spread & u & - & 76.79\% $\pm$ 4.97\% & 0.004 $\pm$ 0.000 & 12.32\% $\pm$ 3.27\% & 10.89\% $\pm$ 3.41\% \\
 & h & - & 72.76\% $\pm$ 5.58\% & 0.024 $\pm$ 0.005 & 12.43\% $\pm$ 4.27\% & 14.81\% $\pm$ 3.81\% \\
 & r & - & 31.68\% $\pm$ 8.56\% & 0.002 $\pm$ 0.000 & 3.66\% $\pm$ 1.62\% & 64.65\% $\pm$ 9.02\% \\
\\
\end{tabular}}
\caption{Summary of AISL, empirical coverage percentage, interval width and directional error rates percentage (miss-high and miss-low) for the three CP methods (both considering all ensembles and mean based), standard deviation method and ensemble mean spread method averaged over the full assimilation window computed against Truth.}
\label{Aisl_Truth}
\end{table}

\subsubsection{Comparison of uncertainty quantification with CP and Traditional methods }
Now, we consider traditional ensemble-based methods, namely the standard deviation interval and the ensemble spread, as baseline approaches to compare with ensemble mean based CP methods. The standard deviation interval is scaled to provide a 90\% confidence level, ensuring comparability with the nominal coverage of CP methods, and the ensemble spread represents the raw variability of ensemble members around the mean and does not provide an explicit probabilistic coverage guarantee. As a result, AISL value for ensemble spread method is not able to be computed.
From (Table~\ref{Aisl_QPEns}), when comparing the mean-based CP methods against the standard deviation interval using QPEns as the reference, all CP approaches demonstrate competitive and, in some cases, improved performance. For u, all three CP methods (SCP, NCP, and CQR) achieve AISL of 0.004$\pm$0.001, slightly higher than the performance of the standard deviation interval (0.005$\pm$0.000) while maintaining coverage close to the nominal level (89.85-90.02\% for CP vs 94.53-95.71\% for ensemble spread and standard deviation methods). Notably, CP methods produce slightly narrower intervals (0.002) compared to the standard deviation and ensemble spread approaches (0.004), indicating sharper uncertainty estimates. For h, CP methods again yield comparable AISL values (0.051-0.055) to the standard deviation interval (0.051$\pm$0.019), while maintaining similar coverage levels (89.88-89.93\% for CP; 88.80\% for standard deviation; 87.13\% for ensemble spread). For variable r, CP methods achieve AISL values of 0.008, slightly higher than the standard deviation interval (0.007$\pm$0.003), but maintain significantly better coverage of 89.81-89.85$\pm$ than standard deviation and ensemble spread methods with 31\% coverage and higher values of miss high (65-66\%).

The same comparison is evaluated against the true model state (Table~\ref{Aisl_Truth}) and for u, all mean-based CP methods (SCP, NCP, and CQR) produce AISL values around 0.010-0.011, which are slightly higher than the standard deviation interval (0.008). However, CP methods show lower coverage (61-64\%) compared to the standard deviation (80.46\%) and ensemble spread (76.79\%), indicating that the QPEns-based CP interval tends to underestimate the true uncertainty for this variable. This is further reflected in the larger directional miss rates for CP (16-20\% miss-high and miss-low) compared to the standard deviation (8-11\%) and ensemble spread (10-12\%), although CP intervals remain slightly sharper in width (0.002 vs 0.004). For h, the performance of CP methods becomes more comparable to the standard deviation interval and ensemble spread. The AISL values for CP (0.066-0.073) are very similar to the standard deviation (0.068$\pm$0.020), with coverage levels of 80-86\% for CP compared to 77.53\% for the standard deviation and least for ensemble spread with 72.76\%. The interval widths are also comparable between methods (0.020-0.027). For variable r, CP methods clearly outperform the standard deviation interval and ensemble spread in terms of coverage. While the standard deviation and ensemble spread approaches shows very poor coverage (31.68-32.42\%), all CP methods maintain significantly higher coverage (89-92\%). The AISL values are comparable (0.009 for CP vs 0.008 for standard deviation), but the large miss-high rate of the standard deviation and ensemble spread (64-65\%) highlights its inability to capture the asymmetric uncertainty in r. In contrast, CP methods maintain much more balanced and smaller miss rates.

\begin{figure}[h!]
\begin{center}
\includegraphics[width=0.45\columnwidth]{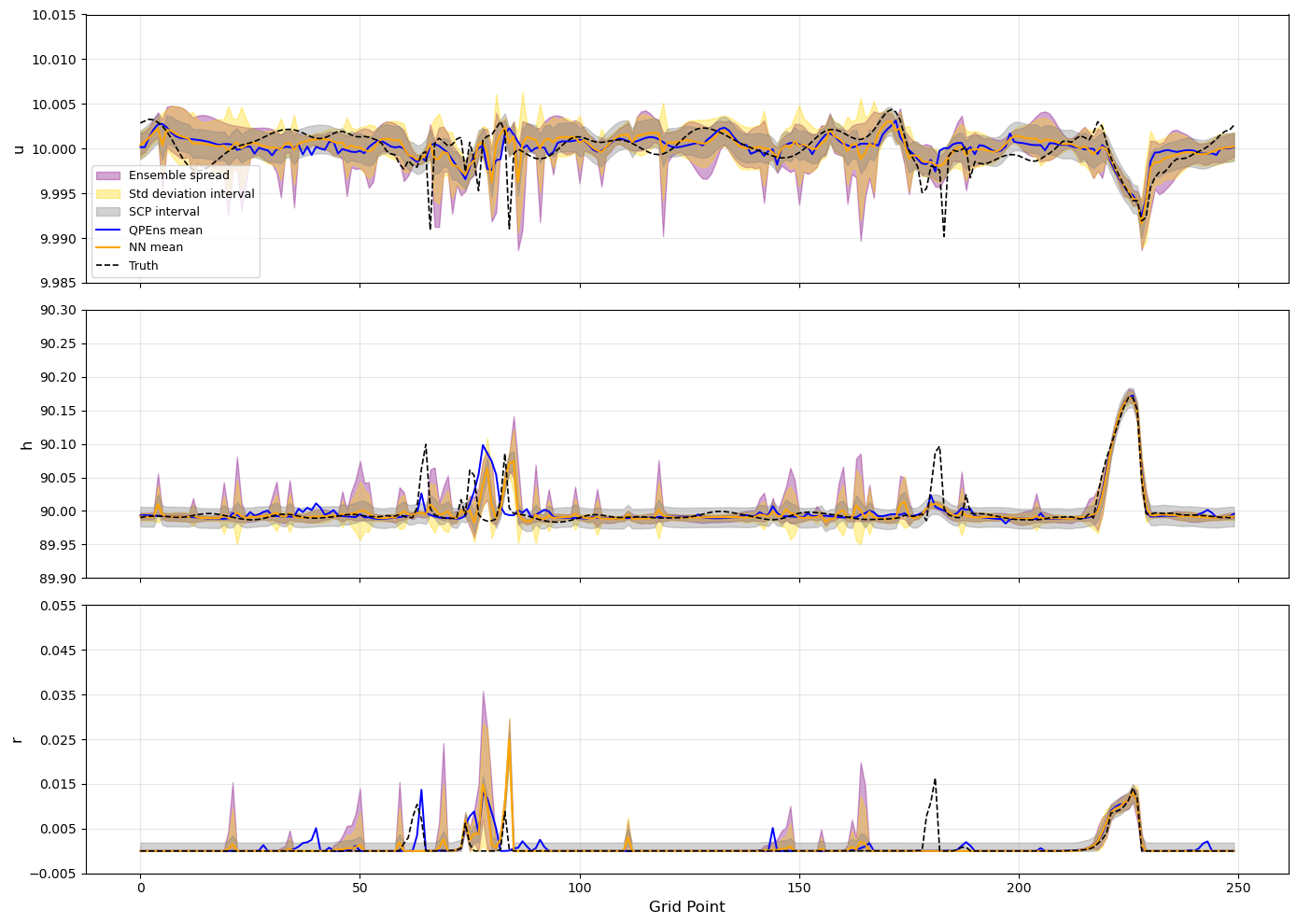}
\includegraphics[width=0.45\columnwidth]{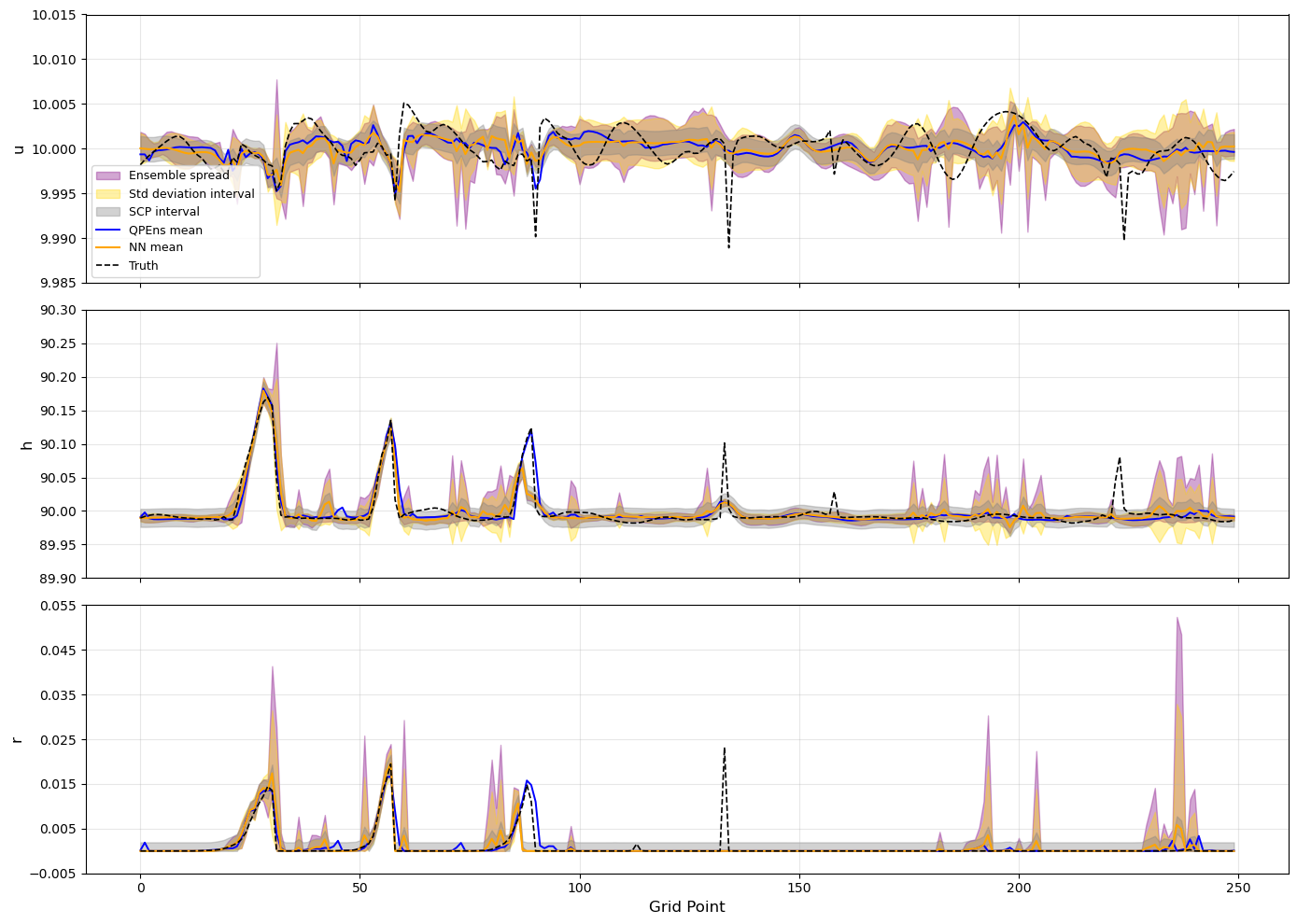}
\caption{Truth (black dashed) and ensemble mean for QPEns (blue) and NN (orange) at timestep 25(left) and 160(right). Gray shading represents uncertainty calculated with mean based SCP Method, std deviation and ensemble spread for the three prognostic variables $u$ (upper), $h$ (middle), $r$ (lower).}
\label{SCP_Trad}
\end{center}

In Figure~\ref{SCP_Trad} for timestep 25 and 160, the SCP mean-based intervals show a consistent balance between sharpness and coverage when compared with traditional methods. For u, the SCP interval closely follows the QPEns mean and remains well aligned with the truth, while maintaining a moderate interval width that is visibly narrower than the standard deviation interval and more stable than the fluctuating ensemble spread. The standard deviation interval tends to be slightly wider and smoother, occasionally overestimating uncertainty, whereas the ensemble spread exhibits larger variability and sporadic spikes.
For h, SCP again produces tight and stable intervals that track both QPEns and truth effectively, particularly around localized peak. In contrast, the standard deviation interval appears slightly wider and smoother, missing some sharp variations, while the ensemble spread shows intermittent bursts of large spread, indicating less consistent uncertainty representation. For variable r, where non-Gaussian behaviour is prominent, SCP maintains controlled and adaptive interval widths, capturing key peaks while avoiding excessive widening. The standard deviation interval, although smooth, fails to fully capture these localized extremes, while the ensemble spread shows pronounced spikes and irregular variability. 

\end{figure}
\begin{figure}[h!]
\begin{center}
\includegraphics[width=0.45\columnwidth]{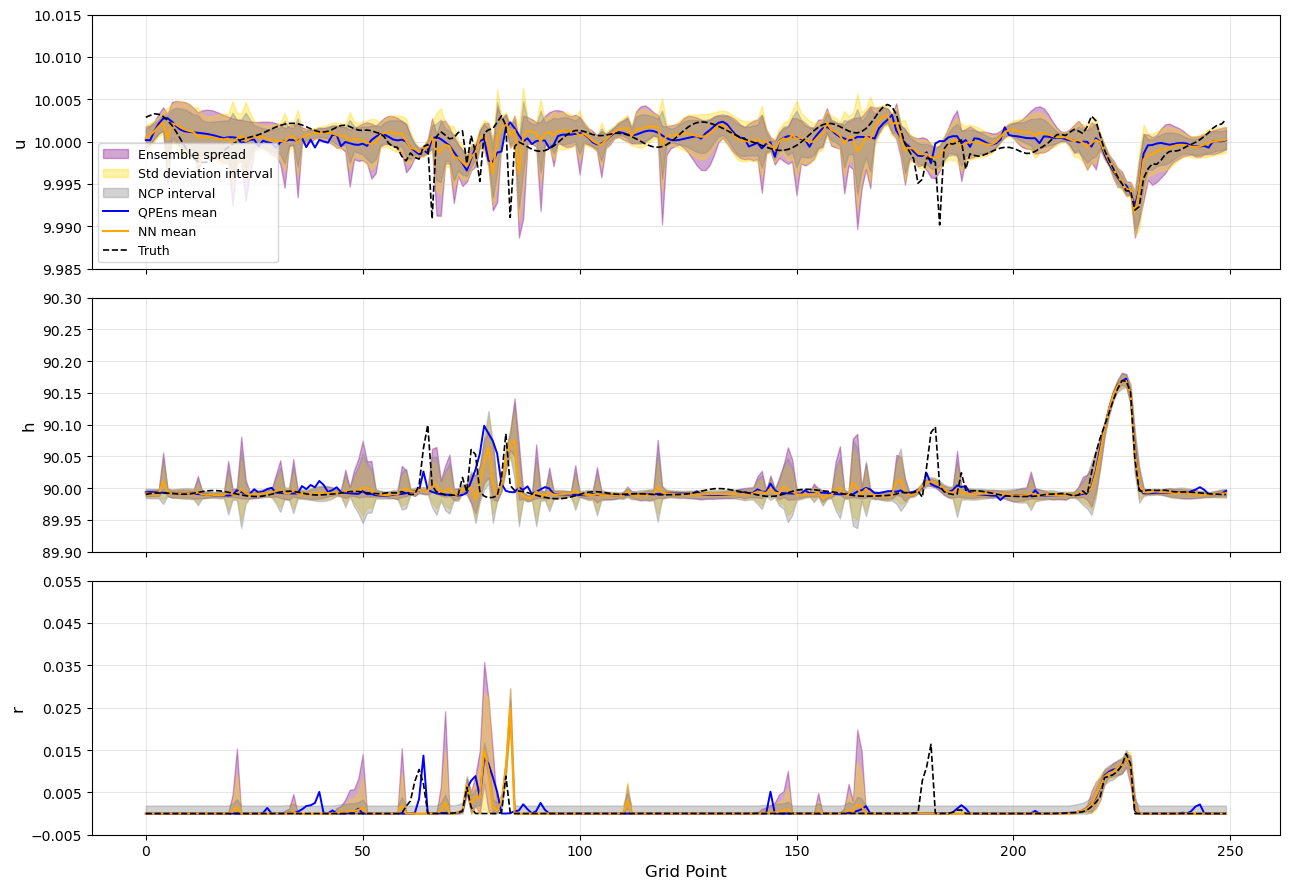}
\includegraphics[width=0.45\columnwidth]{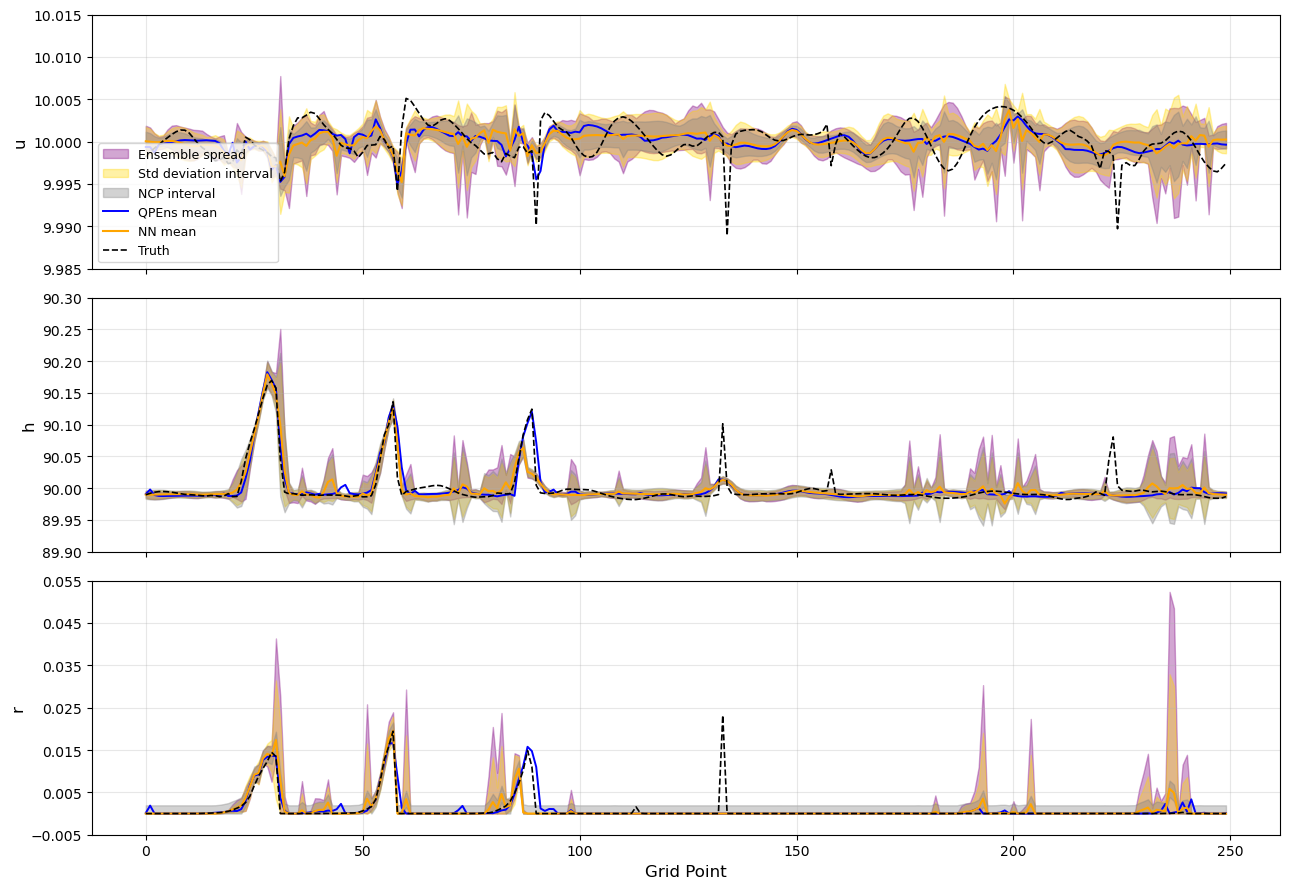}
\caption{Truth (black dashed) and ensemble mean for QPEns (red) and NN (orange) at timestep 25 (left) and 160 (right). Gray shading represents uncertainty calculated with mean based NCP Method, std deviation and ensemble spread for the three prognostic variables $u$ (upper), $h$ (middle), $r$ (lower).}
\label{NCP_Trad}
\end{center}
\end{figure}
Now in Figure~\ref{NCP_Trad}, the NCP (mean-based) intervals exhibit a sharper but more conservative behaviour compared to SCP, particularly in how the interval width adapts relative to QPEns and the truth. For u, NCP produces consistently narrower and smoother intervals than SCP, closely following the QPEns mean but often appearing slightly under covered with respect to the truth, especially in regions with stronger deviations. In contrast, SCP shows slightly wider and more adaptive intervals that better accommodate local variability. The standard deviation interval remains wider than both CP methods, while the ensemble spread shows large fluctuations and instability.
For h, NCP maintains tight and structured intervals that capture the main peaks but are generally narrower than SCP, indicating reduced spread around sharp gradients. The standard deviation interval again appears moderately wider, whereas the ensemble spread exhibits high variability. For variable r, since the same SCP-based formulation is used, both methods produce identical interval structures, showing controlled and adaptive interval widths and the standard deviation interval remains relatively rigid and unable to fully adapt to localized extremes, while the ensemble spread shows strong spikes and irregular variability.
\begin{figure}[h!]
\begin{center}
\includegraphics[width=0.45\columnwidth]{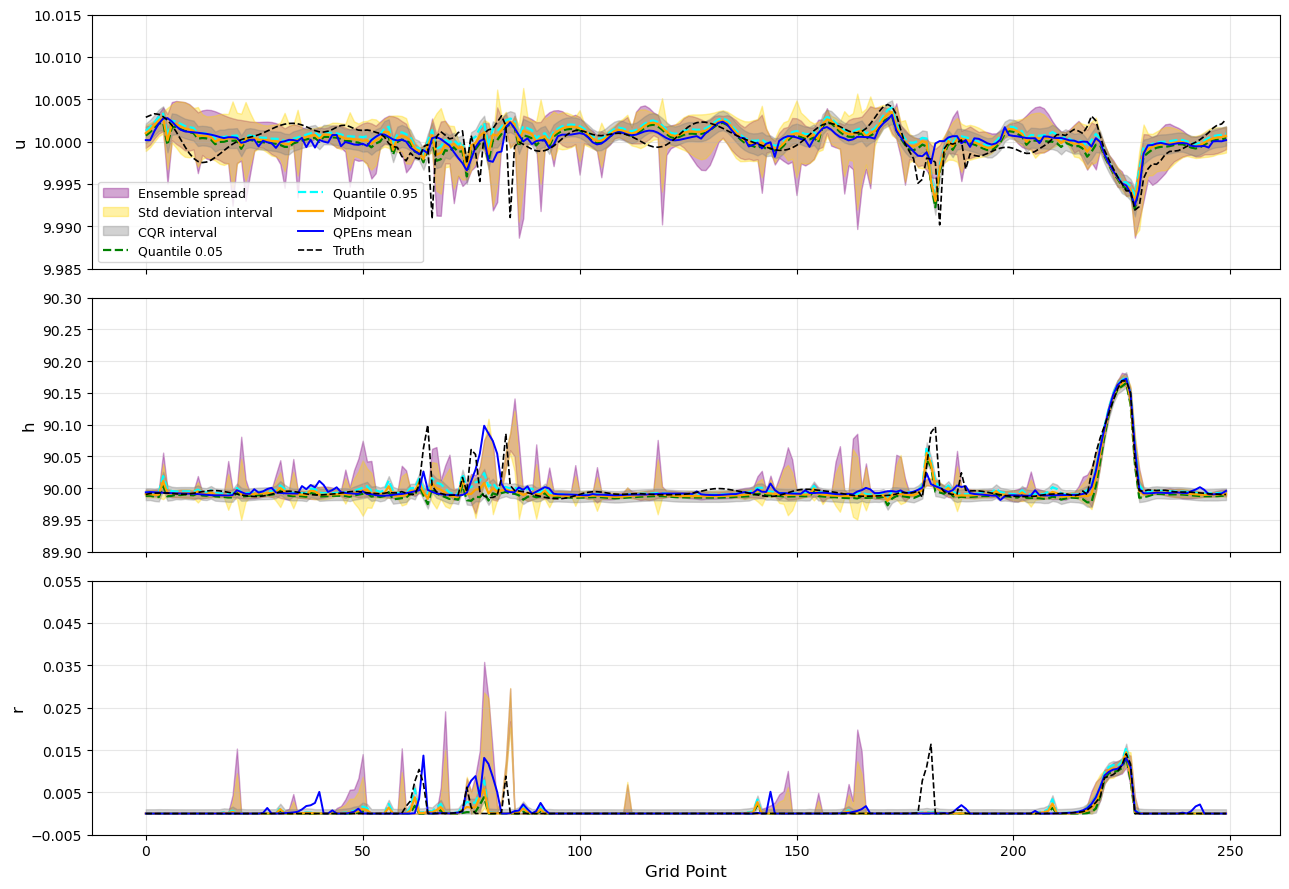}
\includegraphics[width=0.45\columnwidth]{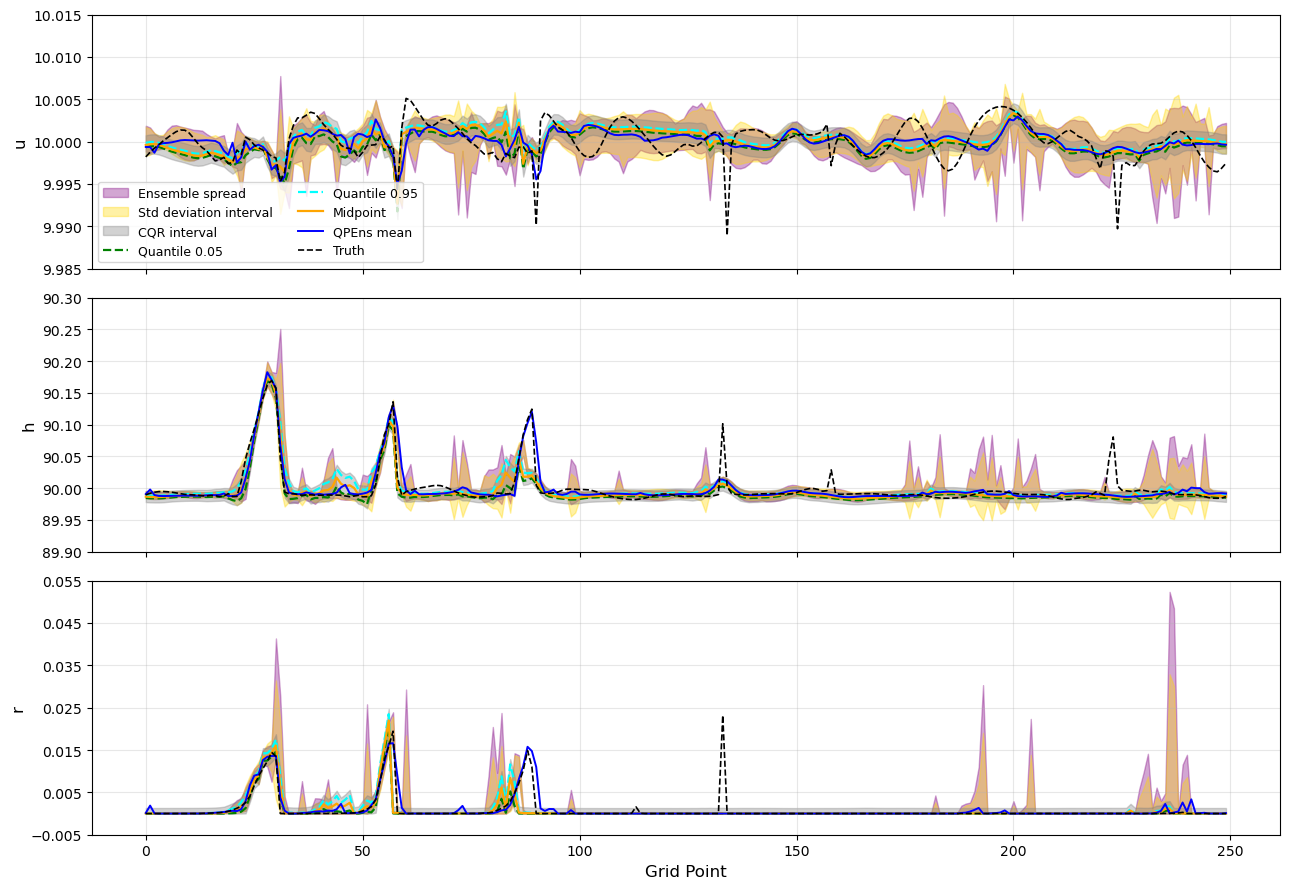}
\caption{Truth (black dashed) and ensemble mean of QPEns (blue),upper quantile ($1-\alpha/2$=0.95,cyan), lower quantile ($\alpha/2$=0.05,green) and midpoint of quantiles (orange) at timestep 25 (left) and 160 (right). Gray shading represents uncertainty calculated with mean based CQR Method, std eviation and ensemble spread for the three prognostic variables $u$ (upper), $h$ (middle), $r$ (lower).}
\label{CQR_Trad}
\end{center}
\end{figure}
In Figure~\ref{CQR_Trad}, the CQR (mean-based) intervals demonstrate a more adaptive and asymmetric representation of uncertainty compared to SCP and NCP, particularly in how the quantile bounds respond to variations relative to QPEns and the truth. For u, CQR produces intervals that are slightly wider than NCP, allowing better representation of local deviations while still remaining smoother than the highly fluctuating ensemble spread. Compared to SCP, CQR shows comparable alignment with QPEns but with more adaptive upper and lower bounds, capturing asymmetry in deviations from the truth. The standard deviation interval remains more symmetric than ensemble spread which continues to show more variability.
For h, CQR effectively captures sharp localized peaks at both timesteps, with the quantile bounds expanding appropriately around strong gradients. The standard deviation interval appears smoother and slightly less flexible to these peaks, while the ensemble spread shows irregular bursts in high variability regions. For variable r, CQR clearly highlights its strength in handling non-Gaussian and skewed behaviour. The quantiles expand asymmetrically around sharp spikes, capturing the structure of the truth more effectively than both SCP and NCP. In contrast, the standard deviation interval remains relatively rigid and fails to adapt to these localized extremes similar to the ensemble spread.

\subsection{Comparison of EnKF and CNN with and without CP Perturbations}
Two configurations are considered for incorporating CP based uncertainty estimates into the data assimilation cycle. In the first configuration, CP perturbations are applied to the CNN analysis, while in the second configuration, CP perturbations are applied to the EnKF analysis (see Figure~\ref{CP&DA}). For each configuration, we evaluate two cases: EnKF and CNN analyses without CP perturbations, and the corresponding EnKF and CNN analyses with CP perturbations applied at the analysis or output stage. For each CP method (SCP, NCP, and CQR) computed based on ensembles, the EnKF and CNN results without CP perturbations are identical since these CP values are not used and any differences between methods therefore arise solely from the inclusion of CP perturbations. RMSE is used to compare the results of each method. RMSE is computed as the mean over 200 experiments for each method. During the first 20 assimilation cycles, all configurations exhibit nearly identical RMSE values, corresponding to the spin-up phase of the data assimilation system.

\begin{figure}[h!]
\begin{center}
\includegraphics[width=0.70\columnwidth]{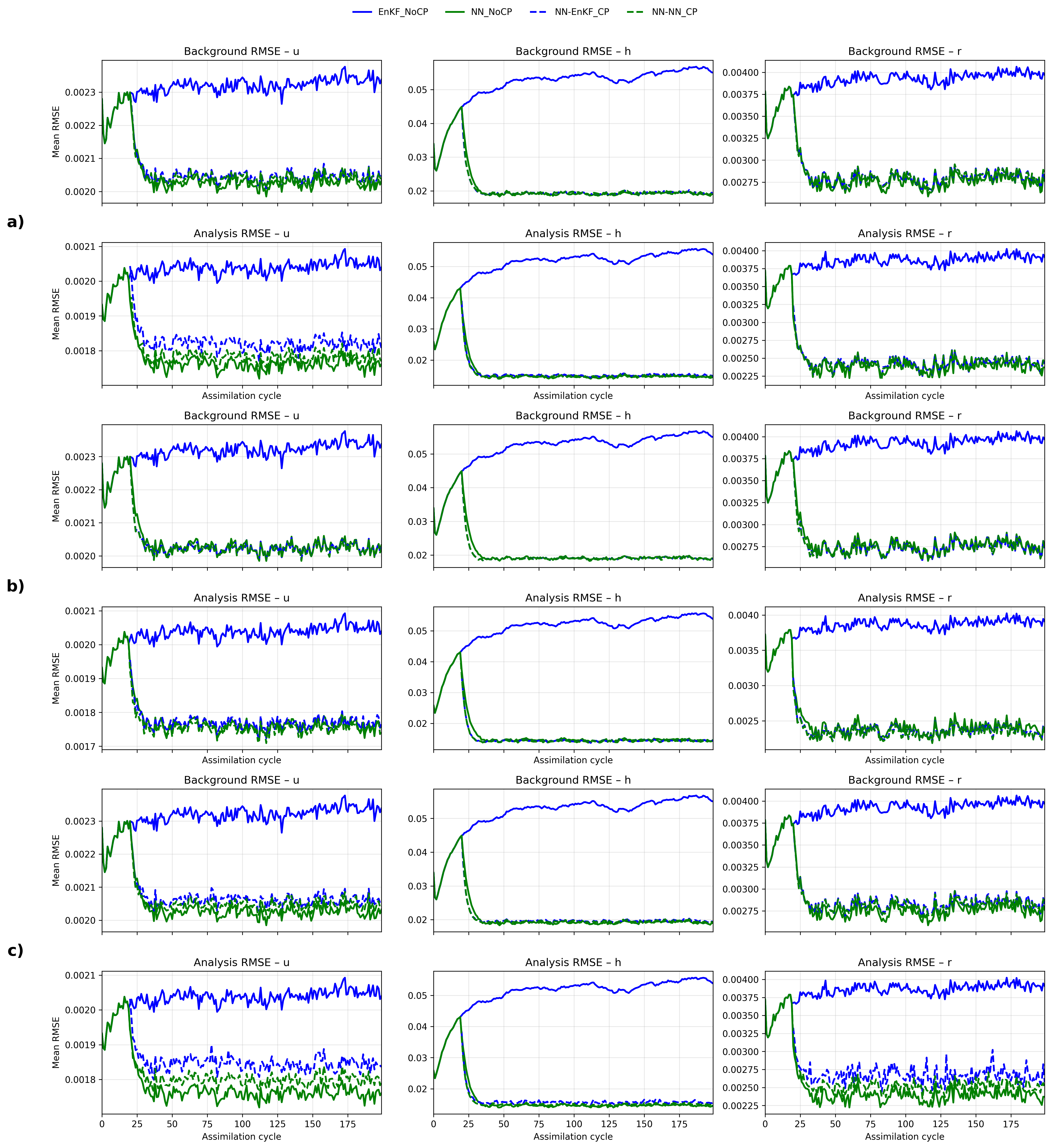}
\caption{RMSE analysis for the variables $u$ (left), $h$ (middle), $r$ (right) of SCP (a), NCP (b) and CQR (c) methods and comparing EnKF and CNN analysis with and without CNN-CP perturbations.}
\label{NN_DA_CP}
\end{center}
\end{figure}

Figure~\ref{NN_DA_CP} shows the RMSE evolution for the background and analysis states when CP perturbations are introduced at the CNN output before the EnKF update from timestep 21. In this configuration, the introduction of CP uncertainty consistently improves the EnKF performance compared to the no-CP baseline. This improvement is reflected in both the background and analysis RMSE, indicating that the perturbations introduced through CP propagate through the assimilation cycle and help maintain a more informative ensemble representation. The background of CP-influenced EnKF and the CP-influenced NN produce are same. This indicates that the background state used by the EnKF directly reflects the CP-perturbed CNN output. 
For all the CP methods, the reduction in RMSE is consistent across all three variables. In both methods, the EnKF with CP perturbations maintain lower RMSE values compared to the no-CP configuration for both background and analysis estimates. In the CQR method, the EnKF continues to show lower RMSE than the no-CP configuration across all variables, but higher in comparison to SCP and NCP methods especially for u and r. The same bias is observed in the SCP method for u but not in NCP which points us to the lack of potential in this setup to capture the stochastic perturbations added to u at each timestep. 
\begin{figure}[h!]
\begin{center}
\includegraphics[width=0.70\columnwidth]{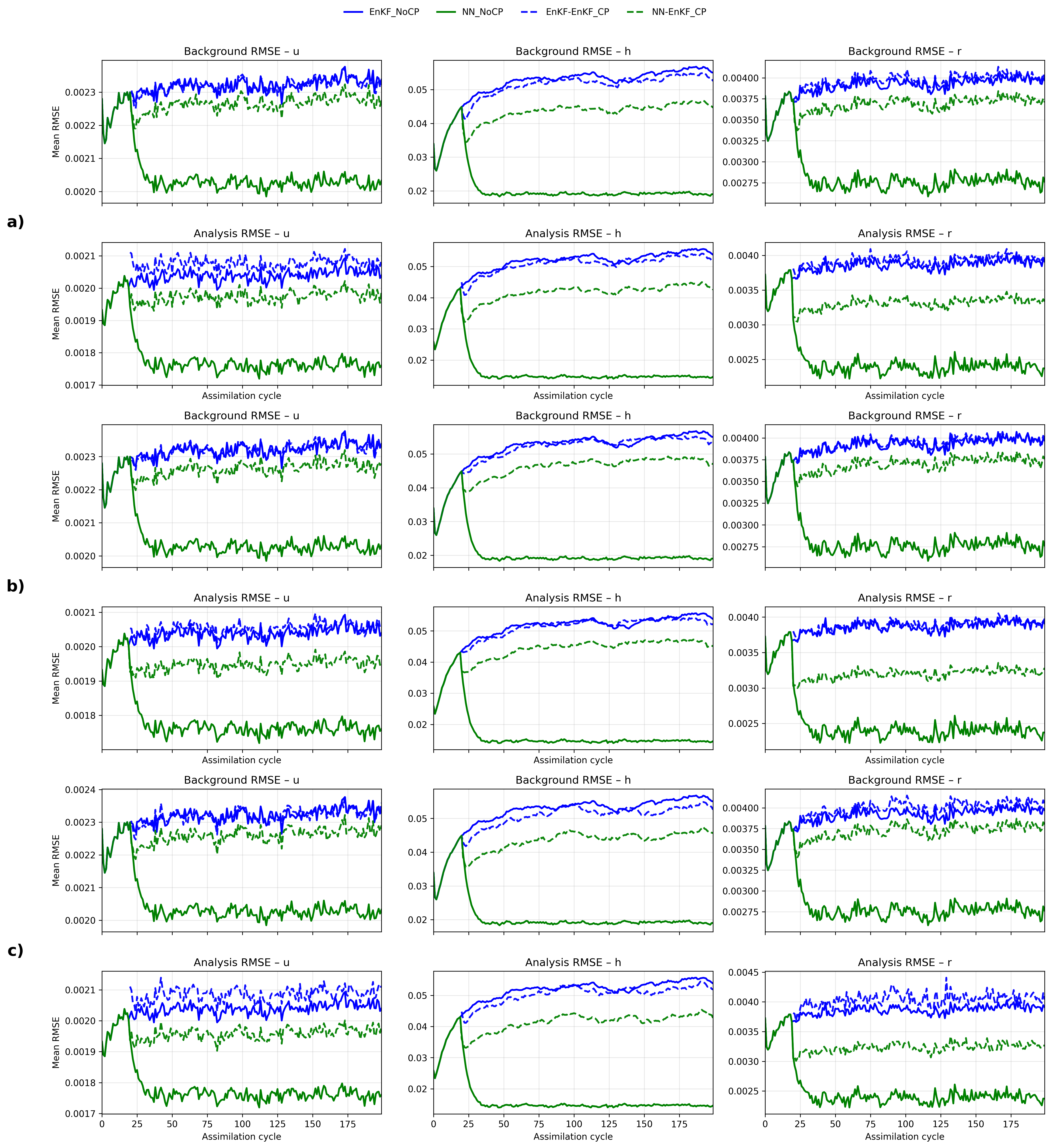}
\caption{RMSE analysis for the variables $u$ (left), $h$ (middle), $r$ (right) of SCP (a), NCP (b) and CQR (c) methods and comparing EnKF and CNN analysis with and without EnKF-CP perturbations.}
\label{EnKF_DA_CP}
\end{center}
\end{figure}

In the second configuration, CP perturbations are introduced directly into the EnKF analysis rather than the CNN output. Figure~\ref{EnKF_DA_CP} shows the corresponding background and analysis RMSE evolution for all variables. In contrast to the first configuration, the CNN output is not used to update the EnKF state when CP perturbations are applied. Instead, the conformal perturbations are added directly to the EnKF analysis, while the CNN output acts only as a post-processing step. As a result, the influence of CP on the EnKF and NN behaviour differs from the previous configuration.
For all the CP methods, introducing CP perturbations into the EnKF analysis does not lead to noticeable improvements in RMSE compared to the no-CP EnKF baseline. The background and analysis RMSE for the EnKF remain very similar to the no-CP configuration for variable h but worse for variable u in SCP and CQR and also for variable r in CQR, indicating that these perturbations do not significantly alter the assimilation performance. However, the NN results in this configuration show slightly higher RMSE compared to the NN without CP. This behaviour arises because the CNN output is not used to influence the EnKF update; instead, it is applied only as a post-processing step. Consequently, the CP perturbations introduced in the EnKF analysis do not improve the NN estimates and can lead to a small degradation in NN performance.
For the CQR method, a different behaviour can be observed. In this case, the perturbations lead to a slight reduction in EnKF analysis RMSE across all variables compared to the no-CP baseline. Although the overall improvements remain decent, the reduction in RMSE is more stable and visible than for SCP and CQR, indicating that NCP interacts more effectively with the EnKF update when perturbations are introduced directly into the analysis.

\section{Concluding remarks}

In this study, we investigated the integration of CP based uncertainty quantification into a hybrid ML-data assimilation framework. Specifically, we evaluated three variants of the CP-SCP, NCP, and CQR. We analyzed their performance through empirical coverage, interval width, AISL scores, and spatial snapshots. The analysis first focused on the intrinsic behaviour of CP methods, including a comparison between ensemble-based and mean-based formulations, with traditional methods and subsequently on their impact when incorporated into the data assimilation cycle under different perturbation configurations. It is also worth noting that, while CP requires exchangeability of the data, a condition not strictly satisfied by our ensemble dataset, the results confirm that CP remains practically valid in this dynamical system setting, producing coverage consistently close to the nominal level.
All CP methods achieve coverage close to the nominal 90\% level when evaluated against the reference distribution, confirming their validity for uncertainty quantification. However, clear differences emerge in terms of stability and adaptivity. The comparison between ensemble-based and mean-based formulations reveals that the optimal representation depends on the CP method: SCP and CQR benefit from using the ensemble mean, leading to tighter and more stable intervals, whereas NCP performs better when applied directly to the ensemble members, as its normalization inherently captures local variability. Across methods, NCP provides the best overall balance between sharpness and reliability, while CQR demonstrates clear advantages in handling non-Gaussian and asymmetric variables, particularly for rainfall. And SCP produces more uniform and less adaptive intervals due to its global conformity threshold.
When compared with traditional uncertainty quantification approaches, namely the standard deviation interval and ensemble spread, CP methods demonstrate clear advantages. While the standard deviation interval assumes Gaussianity and provides fixed symmetric bounds, CP methods produce adaptive intervals that maintain sharpness while achieving more consistent coverage, especially for non-Gaussian variables. The ensemble spread, lacking a probabilistic guarantee, exhibits high variability and cannot provide calibrated uncertainty estimates. Notably, CP methods significantly outperform traditional approaches for skewed variables such as rainfall, where Gaussian assumptions break down.

We further evaluated two perturbation configurations that differ in the placement of CP-based uncertainty within the hybrid framework: perturbations applied at the CNN output and at the EnKF analysis. Introducing CP perturbations at the CNN output consistently improves the EnKF analysis across all methods, with the strongest gains observed for NCP, followed by CQR and SCP. So, applying CP directly at the EnKF analysis stage results in limited or no improvement, indicating that modifying the analysis state after the update is less effective than improving the prior ensemble representation. Also, the CNN itself does not improve under CP perturbations, as it is not retrained to incorporate conformal uncertainty. Overall, the results highlight that both the uncertainty structure (choice of the CP method) and its placement within the assimilation pipeline are critical to enhancing hybrid ML data assimilation systems. Among the approaches tested, the combination of NCP with CNN output perturbation yields the most consistent improvements.

Future work will focus on improving the inclusion of CP into data assimilation cycling by moving from the current offline computation of CP conformity scores to an online framework, where prediction intervals are updated sequentially during data assimilation cycle. Also, we focus on more tightly integrating CP within the data assimilation cycle by retraining the CNN online so that uncertainty estimates evolve consistently with the dynamically updated model state, rather than being applied post hoc. In addition to this, we will focus on analyzing whether incorporating aleatoric and epistemic uncertainty can improve the intervals, producing insights for more reliable and interpretable atmospheric forecasting. By separating inherent data variability from model-related uncertainty, we aim to develop more physically consistent ensemble representations and further improve the performance and reliability of hybrid ML-data assimilation systems.

\par\null

\section*{\texorpdfstring{{Data and
Code}}}
{The source code (https://doi.org/10.5281/zenodo.4354602, Ruckstuhl et al., 2020) includes the necessary scripts to produce the training data and CNN and https://github.com/Catherinegeo98/Uncertainty-quantification-via-conformal-prediction-in-data-assimilation contains the necessary codes for CP anaylsis and perturbing CP estimates into data assimilation.}

\section*{\texorpdfstring{{Acknowledgements}}}
 The authors are grateful for the financial support of the Klaus Tschira Stiftung gGmbH, whose funding made this project possible. The authors also wish to thank Yvonne Ruckstuhl for her valuable insights and for the foundational work established in her prior research, upon which this study builds.

\section*{\texorpdfstring{{Conflict of Interest Statement}}}
 The authors declare that they have no conflict
of interest.

\par\null

\selectlanguage{english}
\bibliography{bibliography/pub.bib%
}

\end{document}